\documentclass{article}

\usepackage[preprint]{neurips_data_2024}

\usepackage[utf8]{inputenc} 
\usepackage[T1]{fontenc}    
\usepackage{hyperref}       
\usepackage{url}            
\usepackage{booktabs}       
\usepackage{amsfonts}       
\usepackage{nicefrac}       
\usepackage{microtype}      
\usepackage[dvipsnames]{xcolor}         

\usepackage{graphicx}
\usepackage{mdframed}
\usepackage{booktabs}
\graphicspath{{imgs/}}
\usepackage{subcaption}
\usepackage{amsmath}
\usepackage{multirow}
\usepackage{makecell}
\usepackage{wrapfig}
\usepackage{tabularx}

\definecolor{ao(english)}{rgb}{0.0, 0.5, 0.0}
\definecolor{azure(colorwheel)}{rgb}{0.0, 0.5, 1.0}

\title{Visual Transformation Telling}

\author{%
    Wanqing Cui\textsuperscript{1,3,}\thanks{These authors contributed equally to this work.} ~~%
    Xin Hong\textsuperscript{4,}\footnotemark[1] ~~%
    Yanyan Lan\textsuperscript{4,}\thanks{Corresponding author.} ~~%
    Liang Pang\textsuperscript{2,3} ~~%
    Jiafeng Guo\textsuperscript{1,3} ~~%
    Xueqi Cheng\textsuperscript{1,3} ~~%
    \\
    {\textsuperscript{1} CAS Key Laboratory of Network Data Science and Technology,} \\{Institute of Computing Technology, Chinese Academy of Sciences, Beijing, China} \\
    {\textsuperscript{2} Data Intelligence System Research Center,} \\
    {Institute of Computing Technology, CAS, Beijing, China} \\
    {\textsuperscript{3} University of Chinese Academy of Sciences, Beijing, China} \\
    {\textsuperscript{4} Institute for AI Industry Research, Tsinghua University, Beijing, China} \\
 {\tt\small \{cuiwanqing18z, pangliang, guojiafeng, cxq\}@ict.ac.cn}\\
 {\tt\small hongxin@air.tsinghua.edu.cn~~lanyanyan@tsinghua.edu.cn}
}

\begin{document}

\maketitle

\begin{abstract}

Humans can naturally reason from superficial state differences (e.g. ground wetness) to transformations descriptions (e.g. raining) according to their life experience.
In this paper, we propose a new visual reasoning task to test this transformation reasoning ability in real-world scenarios, called \textbf{V}isual \textbf{T}ransformation \textbf{T}elling (VTT). Given a series of states (i.e. images), VTT requires to describe the transformation occurring between every two adjacent states. Different from existing visual reasoning tasks that focus on surface state reasoning, the advantage of VTT is that it captures the underlying causes, e.g. actions or events, behind the differences among states. We collect a novel dataset to support the study of transformation reasoning from two existing instructional video datasets, CrossTask and COIN, comprising 13,547 samples. Each sample involves the key state images along with their transformation descriptions. Our dataset covers diverse real-world activities, providing a rich resource for training and evaluation. To construct an initial benchmark for VTT, we test several models, including traditional visual storytelling methods (CST, GLACNet, Densecap) and advanced multimodal large language models (LLaVA v1.5-7B, Qwen-VL-chat, Gemini Pro Vision, GPT-4o, and GPT-4). Experimental results reveal that even state-of-the-art models still face challenges in VTT, highlighting substantial areas for improvement.

  
\end{abstract}

\section{Introduction}
\label{sec:introduction}

\begin{wrapfigure}{r}{0.6\textwidth}

\vspace{-15pt}

\centering
\includegraphics[width=\linewidth]{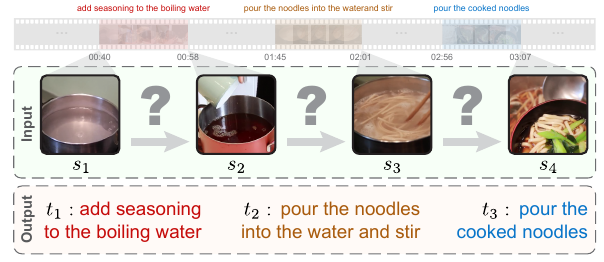}
\caption{
\textbf{Visual Transformation Telling (VTT).} Given \textit{states}, which are images extracted from videos, the goal is to reason and describe \textit{transformations} between every two adjacent states.
}
\label{fig:vtt_illustration}

\vspace{-10pt}

\end{wrapfigure}

What comes to your mind when you are given a series of images, e.g. Figure~\ref{fig:vtt_illustration}? We may first notice the content of each image, then connect them in our mind, and finally conclude a series of events from images, i.e., the entire intermediate process of cooking noodles. In fact, as described in Piaget's theory of cognitive development~\cite{bovetPiagetTheoryCognitive1976,piagetRoleActionDevelopment1977}, this is a typical reasoning process from states (i.e., single images) to transformation (i.e., changes between images).
This ability, perceiving and analyzing transformations between states, marks a significant advancement in cognitive development. In the preoperational stage (2-7 years old), children tend to concentrate on static states and often overlook these dynamic transformations. However, as they enter the concrete operational stage (7-12 years old), their cognitive capabilities evolve, enabling them to gradually appreciate and understand the transformations between states.


Interestingly, the development of computer vision, especially at the stage of deep learning, follows a similar pattern. Early computer vision primarily focused on tasks such as image classification, image detection, image captioning, image question answering, and image generation, aiming to understand or generate static states, and it has achieved satisfactory results. Recent multimodal large language models (MLLMs)~\cite{liuVisualInstructionTuning2023,baiQwenVLVersatileVisionLanguage2023,geminiteamGeminiFamilyHighly2024,openaiGPT4TechnicalReport2024} have further benefited from larger data volumes and more extensive model parameters, achieving even greater breakthroughs. As machines' ability to understand and generate static states approaches or surpasses human levels, researchers have shifted focus to dynamic vision tasks. These include visual storytelling~\cite{ting-haoVisualStorytelling2016}, procedure planning~\cite{changProcedurePlanningInstructional2020}, and video generation~\cite{singerMakeAVideoTexttoVideoGeneration2022,hoImagenVideoHigh2022,hongCogVideoLargescalePretraining2022}. However, current models often struggle to understand transformation properly, which leads to mistakes in understanding or creating visual content. For instance, while Sora~\cite{liuSoraReviewBackground2024} can make good videos, she still finds it hard to show simple transformations accurately, like how glass breaks, that she might show water suddenly on the table before the glass has broken. This happens because she can't figure out how the cup on the table turns into spilled water. So, it's crucial to model transformation accurately to handle more complex tasks.

In this paper, we propose a new task, called \textbf{V}isual \textbf{T}ransformation \textbf{T}elling (VTT), to directly evaluate the ability of transformation modeling in real scenarios. VTT task asks models to generate language sentences to describe the transformation for a given series of states, i.e.~images. Different from traditional visual reasoning tasks that only consider state differences, VTT focuses on digging for underlying transformation behind observation. As the images $s_3, s_4$ shown in Figure~\ref{fig:vtt_illustration}, the change in the position of noodles is merely a surface phenomenon, the more fundamental reason is that someone pouring out the noodles, leading to the state transition. Previously, there have been some preliminary studies~\cite{parkRobustChangeCaptioning2019,hongTransformationDrivenVisual2021,qiuGraphRepresentationOrderaware2023a} on transformation. However, they are defined in an artificial environment with extremely simple transformations, which is difficult to simulate the diversity and complexity of transformations in reality. 
In contrast, our data is collected from two real instructional video datasets, including CrossTask\cite{zhukovCrossTaskWeaklySupervised2019} and COIN~\cite{tangCOINLargescaleDataset2019,tangComprehensiveInstructionalVideo2021}, covering diverse daily activities. Both datasets were originally used for evaluating step localization, action segmentation, and other video analysis tasks. Therefore, the main steps required to accomplish a certain job were annotated, including temporal boundaries and text descriptions. We adapt the data based on these labels to suit the VTT task. Specifically, we extracted key images from the video as states for inputs, and directly used their text labels of the main steps as target transformation descriptions.

We benchmark existing models on VTT tasks and conduct extensive analysis. Specifically, considering the similarity of visual storytelling to VTT, we modify typical methods including CST~\cite{gonzalez-ricoContextualizeShowTell2018}, GLACNet~\cite{kimGLACNetGLocal2019}, and Densecap~\cite{johnsonDenseCapFullyConvolutional2016a}. We also test several multimodel large language models (MLLMs), including open source models, i.e., LLaVA v1.5-7B~\cite{liuVisualInstructionTuning2023}, Qwen-VL-chat~\cite{baiQwenVLVersatileVisionLanguage2023}, and close source models, i.e., Gemini Pro Vision~\cite{geminiteamGeminiFamilyHighly2024}, GPT-4~\cite{openaiGPT4TechnicalReport2024} and GPT-4o~\cite{HelloGPT4o}. Experimental results indicate that existing models still have significant scope for improvement, whether models specifically designed for telling or MLLMs trained on large-scale datasets with massive parameters. According to the human evaluation, event the best performing model, i.e. Gemini-1.5, only achieves $3.95$, and $4.17$ (out of 5) on Relevance, and Logical Soundnes, respectively. We qualitatively analyze testing cases and summarize four types of common errors for MLLMs, including bias, misidentification, hallucination, and illogicality. We further explore strategies to improve existing model and find tuning MLLMs on VTT data. We find that tuning MLLMs on VTT data significantly enhances its relevance and logic consistence, indicating that the current training data lacks sufficient information about transformations reasoning. A major future direction will be to construct and provide the model with more data contains transformation information. We also find prompt strategies like forcing the model to predict the overall transformation topic can improve the performance. In addition, explicitly modeling the differences between the states has shown substantial improvement in traditional models, but applying similar modeling to MLLMs is not trivial, which becomes a potential direction for future study.



The contributions of this study can be summarized as follows:
1) We introduce a novel visual transformation telling task and collect a dataset to resolve the limitations of the reasoning of transformation in real-world scenarios. 
2) We benchmark several models, including traditional models and MLLMs (both open-source and closed-source), highlighting the significant room for improvement.
3) We summarize the error types of existing models and provide potential directions for future research. Our code is released at \url{https://github.com/hughplay/VTT}.

\section{Related Works}

Visual reasoning has been considered as one of the next north star of computer vision~\cite{fei-feiSearchingComputerVision2022a}, and is constantly being examined by the new multimodal large models that have emerged in recent years. Early visual reasoning tasks mainly focus on state-level reasoning. CLEVR~\cite{johnsonCLEVRDiagnosticDataset2017} and GQA~\cite{hudsonGQANewDataset2019} concentrate on object relation and logical reasoning. RAVEN~\cite{zhangRAVENDatasetRelational2019a} and V-PROM~\cite{teneyVPROMBenchmarkVisual2020} concentrate on the induction and reasoning of graphic patterns. VCR~\cite{zellersRecognitionCognitionVisual2019} and Sherlock~\cite{hesselAbductionSherlockHolmes2022a} test the machine's ability to learn commonsense knowledge to answer daily questions. In addition to these tasks, there is a series of works related to dynamic reasoning.  Physical reasoning~\cite{melnikBenchmarksPhysicalReasoning2023} evaluates the ability to learn physical rules from data to answer questions or solve puzzles. VisualCOMET~\cite{parkVisualCOMETReasoningDynamic2020} requires reasoning beyond the given state to answer what happened before and will happen next. Visual storytelling~\cite{parkVisualCOMETReasoningDynamic2020} requires logically telling a story from information-incomplete states. The field of visual reasoning tends to shift from static scenes to dynamic ones. While reasoning in dynamic scenes, state and transformation are both crucial, we focus on transformation reasoning to better evaluate and improve this ability, which distinguishes VTT from state-only and more complex composite tasks.

To the best of our knowledge, there are few studies on designing specific tasks for visual transformation reasoning. TVR~\cite{hongTransformationDrivenVisual2021} and OVT~\cite{qiuGraphRepresentationOrderaware2023a} require to predict a sequence of property (e.g. color) changes given the initial and final states. However, the synthetic scenario used in both datasets is far from reality and the property changes are not commonly used to describe transformations in real life. In contrast, VTT emphasizes event-level description, which is a more natural way of describing transformations. Visual storytelling~\cite{ting-haoVisualStorytelling2016,raviAESOPAbstractEncoding2021a} indeed requires event-level description, but transformations are mixed throughout the story, making it difficult to evaluate transformation reasoning specifically. Visual abductive reasoning~\cite{liangVisualAbductiveReasoning2022} has a similar core idea to VTT, which is to find the most likely explanation for incomplete observations. However, VTT aims to reason multiple logically related transformations from states, while their task only requires reasoning a single missing transformation from multiple transformations. Procedure planning~\cite{changProcedurePlanningInstructional2020} aims to complete a job given states, while VTT focuses on explaining transformations between states, which has wider scenarios, such as explaining the wet ground with rain. Furthermore, the requirement for natural language generation in VTT leads to different evaluations and unique challenges, such as generalization on language compositions and transformation combinations. Finally, walkthrough planning~\cite{changProcedurePlanningInstructional2020} has a different target, which is to predict intermediate states.

Another topic related to VTT is visual description. Tasks that describe a single image include image captioning~\cite{farhadiEveryPictureTells2010, kulkarniBabytalkUnderstanding2011}, dense image captioning~\cite{johnsonDenseCapFullyConvolutional2016a}, and image paragraphing~\cite{krauseHierarchicalApproachGenerating2017}, which vary in the level of detail required. Tasks that describe videos include video description~\cite{venugopalanTranslatingVideosNatural2015}, video paragraph description~\cite{yuVideoParagraphCaptioning2016}, grounded video description~\cite{zhouGroundedVideoDescription2019}, dense video captioning~\cite{krishnaDenseCaptioningEventsVideos2017}, and video timeline modeling~\cite{liuVideoTimelineModeling2023a} start to describe events rather than a single state. For example, dense video captioning asks to predict temporal boundaries of key events and describe them. However, these tasks do not explicitly require reasoning about transformations since they provide the full process of transformation throughout frames.

\section{Visual Transformation Telling Dataset}

\subsection{Task Definition}

Visual transformation telling aims to test machines' ability to reason and describe transformations from a sequence of visual states, i.e., images. Formally, $N+1$ images $S=\{s_n\}_{n=1}^{N+1}$ are provided, which are \textit{logically related} and \textit{semantically distinct}. Logically related means these images are associated with a particular event and are arranged in time sequence. 
Semantically different means that adjacent images come from two discontinuous time points and the content they contain has substantially changed, i.e., a transformation.
The objective is then to reason $N$ transformations $T=\{t_n\}_{n=1}^{N}$ between every two adjacent images and describe them in natural language, such that $s_1 \rightarrow t_1 \rightarrow s_2 \rightarrow \cdots \rightarrow t_n \rightarrow s_{n+1} $ is logically sound.

\begin{figure}[t]
     \centering
     \begin{subfigure}[b]{0.38\textwidth}
         \centering
         \includegraphics[width=\textwidth]{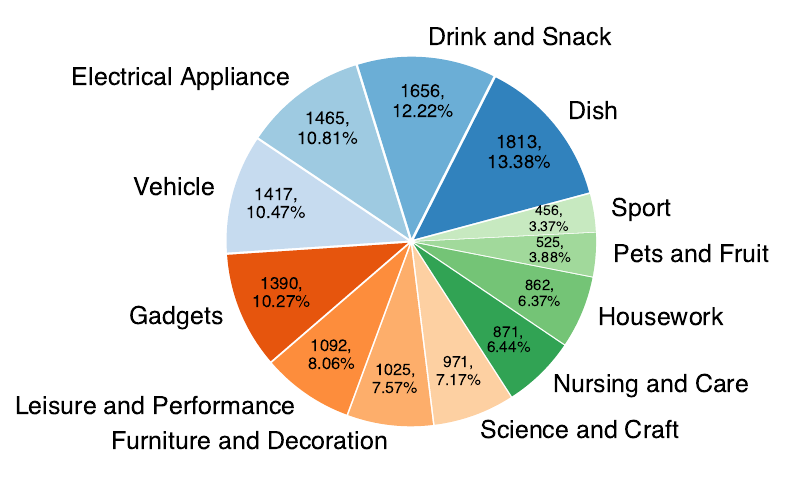}
         \caption{}
         \label{fig:categories_dist}
     \end{subfigure}
     \hfill
     \begin{subfigure}[b]{0.3\textwidth}
         \centering
         \includegraphics[width=\textwidth]{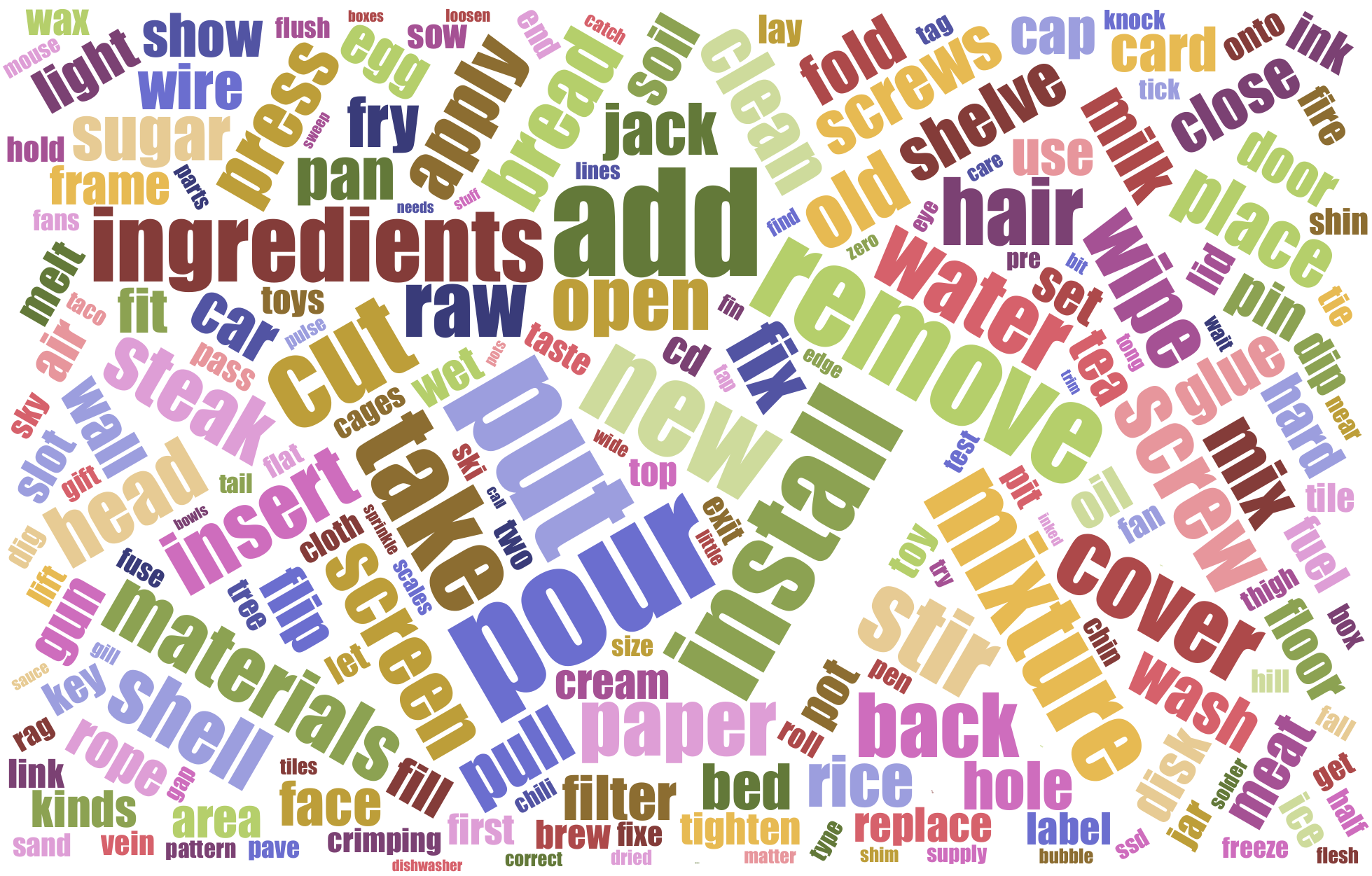}
         \vspace{0.3pt}
         \caption{}
         \label{fig:worldcloud}
     \end{subfigure}
    \hfill
     \begin{subfigure}[b]{0.3\textwidth}
         \centering
         \includegraphics[width=\textwidth]{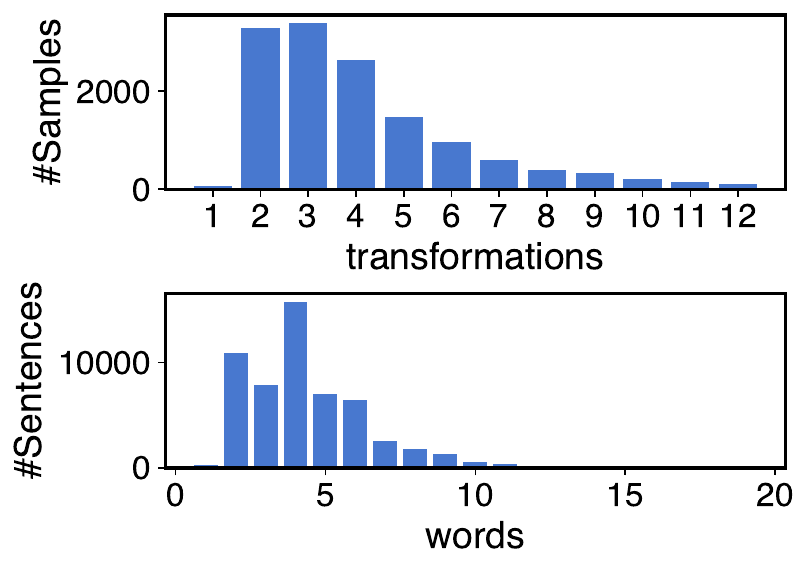}
         \caption{}
         \label{fig:steps_dist}
     \end{subfigure}
     
     \begin{subfigure}[b]{\textwidth}
         \centering
         \includegraphics[width=\textwidth]{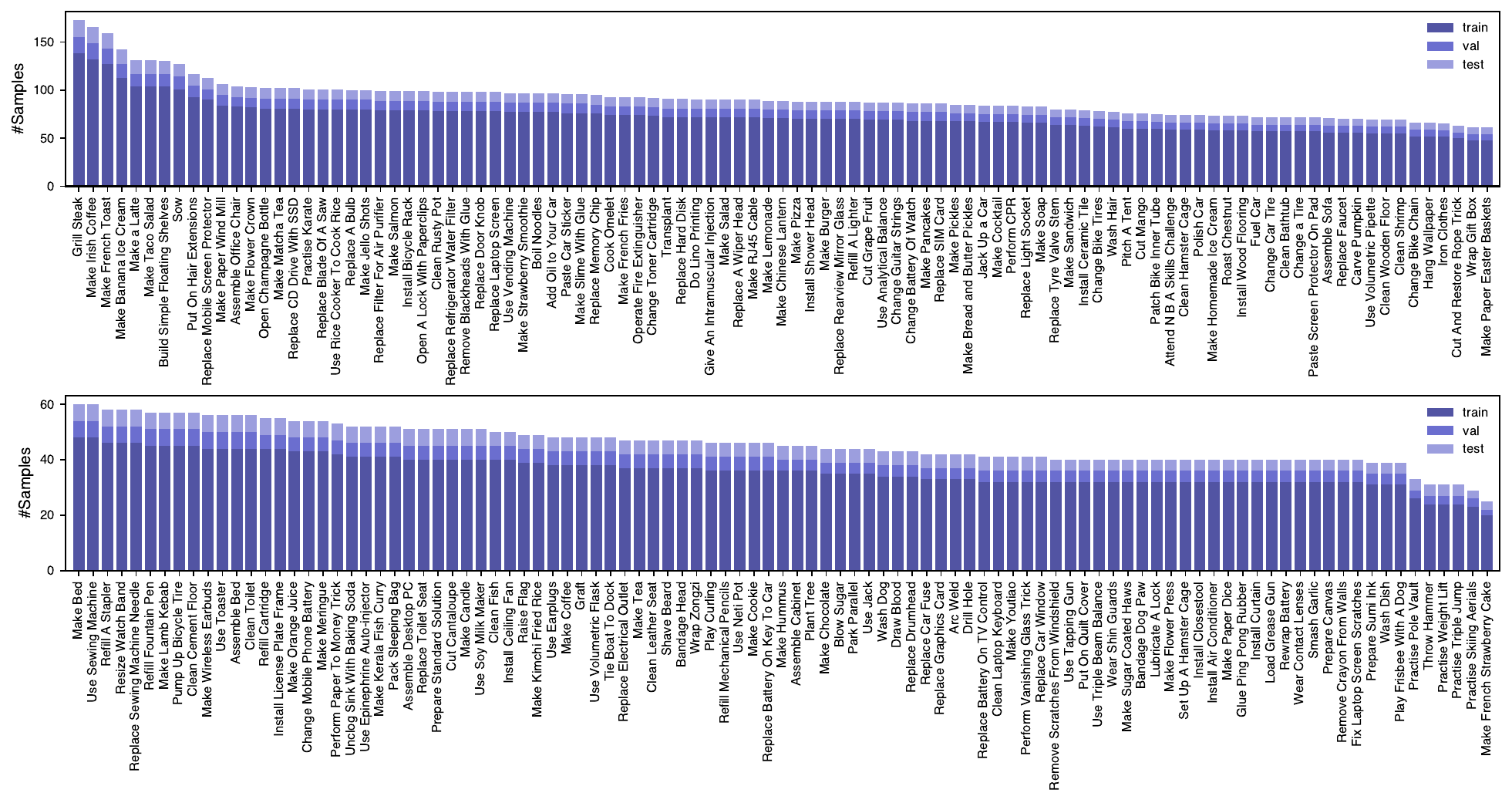}
         \caption{}
         \label{fig:topic_dist}
     \end{subfigure}

    \caption{Distributions of VTT samples. (a) Category. (b) Words. (c) Transformation length (top), sentence length (bottom). (d) Topic.}
    
    \label{fig:distributions}
\end{figure}

\subsection{VTT Dataset Construction}
\label{dataset_construction}

\textbf{Data collection.}
To create a comprehensive dataset of real-world transformations, we chose instructional videos due to their detailed depiction of everyday activities. Specifically, we used two well-known public instructional video datasets: CrossTask~\cite{zhukovCrossTaskWeaklySupervised2019} and COIN~\cite{tangCOINLargescaleDataset2019, tangComprehensiveInstructionalVideo2021}. These datasets provided a rich source of data for our VTT dataset.

\textbf{State and transformation description.}
Figure~\ref{fig:vtt_illustration} illustrates an instructional video from COIN on cooking noodles and how we transformed their annotation into our VTT dataset. We can see that the video is segmented into multiple main steps, each annotated with precise temporal boundaries and text labels.
For \textbf{state image extraction}, the best choice is the frame just before or after a transformation. CrossTask's and COIN's precise temporal segment annotations, which undergo three rounds of refinement~\cite{tangCOINLargescaleDataset2019}, can satisfy this requirement. For the first transformation, we used the first frame of the corresponding step segment as its start state and the last frame as its end state. For the remaining transformations, the end state is extracted in the same way, while the start state shares the end state of the previous transformation. 
For \textbf{transformation descriptions}, we directly used original text labels as transformation labels.
We manually checked the quality of 200 random samples and found that transformations could be reasoned out from states most of the time.
Using this method, we collected 13,547 samples with 55,482 transformation descriptions from CrossTask and COIN, forming our new data for VTT. 

\textbf{Category and topic labels.}
The VTT dataset also includes annotations such as \textit{category}, \textit{topic}, and \textit{transformation description}, which are collected and organized from CrossTask and COIN. Step labels and corresponding segments are provided by both datasets. In CrossTask, step labels were derived from WikiHow, whereas COIN employed experts to define them. Annotators were then tasked with labeling the step categories and corresponding segments for each video. We collected and organized these annotations in a uniform format for the VTT dataset. Both CrossTask and COIN provide topic information, which pertains to the task to be solved. COIN also provides categories as domain information, which are absent in CrossTask. We manually classify all topics from CrossTask into existing categories. Table~\ref{tab:topic_list} in Appendix shows the full list of 12 categories and 198 topics.

\begin{wraptable}{r}{0.6\textwidth}


\vspace{-10pt}
\caption{VTT dataset statistics.}
\label{tab:statistics}
\centering
\small
\setlength{\tabcolsep}{3pt}

\begin{tabular}{lrrrrrr}
\toprule
 & CrossTask & COIN & Train & Val & Test & Total \\
\midrule
Categories & 4 & 12 & 12 & 12 & 12 & 12 \\
Topics & 18 & 180 & 198 & 198 & 198 & 198 \\
Samples & 1825 & 11722 & 10759 & 1352 & 1436 & 13547 \\
States & 12860 & 56169 & 54716 & 6974 & 7339 & 69029 \\
Trans. & 11035 & 44447 & 43957 & 5622 & 5903 & 55482 \\
Unique Trans. & 105 & 749 & 853 & 812 & 806 & 853 \\
\bottomrule
\end{tabular}


\vspace{-8pt}
\end{wraptable}

\textbf{Dataset Split and Statistics.} We split the data randomly into Train / Val / Test sets with samples of 10759 / 1352 / 1436 in the level of topic. 
The detailed topic distribution is shown in Figure~\ref{fig:topic_dist}, and we can see that about half of the topics have over 100 samples. We also summarize the main statistics of the VTT dataset in Table~\ref{tab:statistics}. VTT also requires the model to have the ability to generalize to handle transformation combinations that are not present in the training set.
Figure~\ref{fig:distributions} shows the distribution of the sample categories, keywords, transformation length, and sentence length of VTT. From the category distribution and the word cloud, we can see that the VTT data covers a wide range of daily activities, like dish, electrical application, gadgets, etc. Furthermore, the distribution of transformation length shows diversity, with most samples containing about 2-5 transformations. The average sentence length is around 2-6, indicating that short descriptions make up the majority. In addition to state images and transformation descriptions, each sample in VTT also comes with coarse-grained category labels (e.g., dish) and fine-grained topic labels (e.g., boil noodles).

    





\section{Benchmark on VTT}




\subsection{Model Selection} 
\label{sec:model_selection}
\textbf{Traditional models.}
We first adapt two classic visual story telling methods for comparison, including CST~\cite{gonzalez-ricoContextualizeShowTell2018} and GLACNet~\cite{kimGLACNetGLocal2019}, which are both winners of the visual storytelling challenge~\cite{mitchellProceedingsFirstWorkshop2018}. This is because visual storytelling generates $N$ descriptions from $N$ images, that is similar to our VTT task. 
In addition, we also compared with a dense vieeo captioning method called DenseCap~\cite{johnsonDenseCapFullyConvolutional2016a}, since dense video captioning also has a similar visual description target, which aims to describe a series of events in a video and requires predicting temporal boundaries for events. 
All methods were closely implemented as per the original paper. For a better image understanding, we also provided baseline models with CLIP as image encoder marked with `*'. The implementation details of TTNet as well as the baseline models are described in the supplementary.

\textbf{Multimodal language models.} MLLMs have shown promising capabilities on various vision language benchmarks. To test how well they perform on VTT, we test two open-source models, including LLaVA v1.5-7B~\cite{liuVisualInstructionTuning2023}, Qwen-VL-chat~\cite{baiQwenVLVersatileVisionLanguage2023}. We also test four proprietary vision language models through their public API, including Gemini Pro Vision~\cite{geminiteamGeminiFamilyHighly2024}, GPT-4~\cite{openaiGPT4TechnicalReport2024}, and GPT-4o~\cite{HelloGPT4o}.  Considering that these models may not be well adapted to the task form of VTT, such as language style, differences in word usage, etc., we also tune the LLaVA model with LORA~\cite{Hu2021LoRALA} on VTT for testing.


\subsection{Evaluation Protocol} 

\textbf{Automated metrics.} We follow previous works on visual descriptions~\cite{ting-haoVisualStorytelling2016,krishnaDenseCaptioningEventsVideos2017,liangVisualAbductiveReasoning2022}, and select common used metrics for evaluation, including BLEU@4~\cite{papineniBLEUMethodAutomatic2002}, CIDEr~\cite{vedantamCIDErConsensusbasedimage2015}, METEOR~\cite{banerjeeMETEORAutomaticMetric2005}, ROUGE-L~\cite{linManualautomaticevaluation2002}, SPICE~\cite{andersonSPICESemanticPropositional2016}, and BERT-Score~\cite{zhangBERTScoreEvaluatingText2020}, 

\textbf{Human evaluation.} While we aimed to evaluate the logical consistency of generated transformation descriptions, automatic evaluation of content logical consistency remains a very challenging problem in the NLP field with no solution to date. Generally, only manual evaluation can be performed. Therefore, we asked 25 human annotators to assess the quality of transformation descriptions using a Likert scale ranging from 1 to 5 based on the following criteria: \textit{fluency}, measuring how well-written the transformation is; \textit{relevance}, assessing how relevant the transformations are to the image states; and \textit{logical soundness}, evaluating how well the overall logic conforms to common sense.

\section{Experimental Results and Analysis} \label{sec:experiment}

In this section, we first summarize the ability of various models on VTT and analyze the performance of MLLMs on different data types. Then, we analyze the error types made by the most advanced MLLMs. Finally, we improve the existing model to preliminarily explore how to model visual transformations better, hoping to inspire future study.

\begin{table}[t]
\centering
\small
\caption{Results on VTT evaluated using B@4(BLEU@4), M(METEOR), R(ROUGE-L), C(CIDEr), S(SPICE), BS(BERT-Score), Flu.(Fluency), Rel.(Relevance), and Logic.(Logical Soundness). * indicates changing the image encoder to CLIP. 
`Sep' means inputting each image separately.}
\label{tab:baseline}
\begin{tabular}{l|cccccc|ccc}
\toprule
Model & B@4 & M & R & C & S & BS & Flu. & Rel. & Logic. \\
\midrule
CST & 10.09 & 11.39 & 25.98 & 43.22 & 9.28 & 16.30 & - & - & - \\
CST* & 13.96 & 19.21 & 38.11 & 84.60 & 21.85 & 25.66 & 2.04 & 3.16 & 2.96  \\
GLACNet & 42.77 & 45.26 & 52.98 & 381.48 & 45.33 & 60.12 & - & - & -  \\
GLACNet* & 55.24 & 59.48 & 66.25 & 508.18 & 60.21 & 71.13 & 4.75 & 3.82 & 3.78 \\
DenseCap* & 48.25 & 52.00 & 59.79 & 439.68 & 53.73 & 66.30 & 4.74 & 3.67 & 3.59 \\
\midrule
GPT-4 & 4.73 & 6.74 & 11.76 & 28.24 & 11.66 & 25.84 & - & - & -  \\
GPT-4o & 4.84 & 6.91 & 12.03 & 29.69 & 13.01 & 28.38 & - & - & -  \\
Gemini-1.0 & 8.36 & 10.25 & 19.82 & 47.79 & 16.13 & 31.43 & - & - & -  \\
Gemini-1.5 & 8.51 & 11.1 & 20.62 & 52.25 & 17.93 & 33.88 & 4.95 & 3.95 & \textbf{4.17} \\
Qwen-VL-chat & 4.71 & 4.57 & 10.62 & 15.32 & 6.25 & 23.93 & - & - & -  \\
Qwen-VL-chat (Sep) & 4.70 & 5.62 & 11.23 & 21.91 & 9.38 & 25.64 & - & - & -  \\
LLaVA-1.5-7B & 3.06 & 3.30 & 7.19 & 12.04 & 5.18 & 23.21 & - & - & -  \\
LLaVA-1.5-7B+Topic & 3.14 & 3.46 & 7.56 & 12.49 & 5.95 & 23.76 & 4.79 & 2.08 & 3.07 \\
LLaVA-1.5-7B$_\text{LORA}$ & 31.43 & 32.37 & 40.38 & 268.59 & 33.17 & 49.08 & - & - & -  \\
LLaVA-1.5-7B$_\text{LORA}$+Topic & 33.58 & 34.25 & 41.93 & 289.14 & 35.29 & 50.46 & \textbf{4.98} & 3.10 & 3.76 \\
\midrule
TTNet$_\text{Base}$ & 55.68 & 60.47 & 67.05 & 515.12 & 61.45 & 72.22 & 4.79 & 4.04 & 3.95 \\
TTNet & \textbf{61.22} & \textbf{66.31} & \textbf{71.84} & \textbf{570.63} & \textbf{66.20} & \textbf{76.25} & 4.78 & \textbf{4.10} & 4.11 \\
\bottomrule
\end{tabular}
\end{table}

\begin{figure}[t]
\centering
\includegraphics[width=\linewidth]{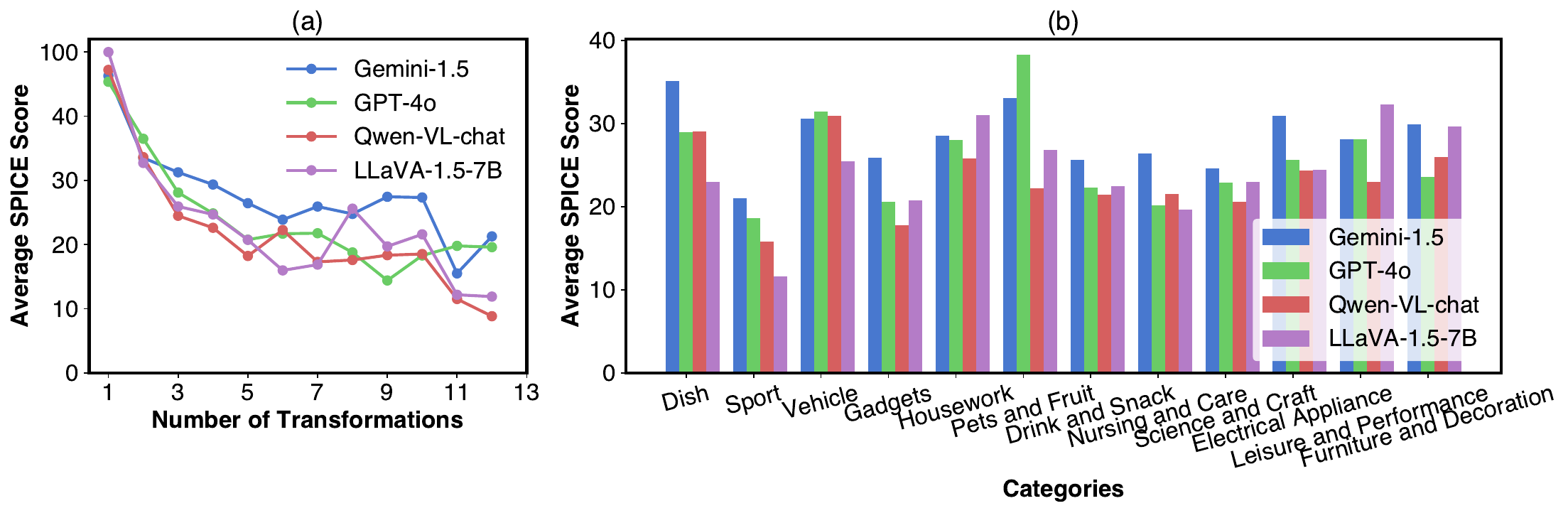}
\caption{Performance of models under different data: (a) The SPICE values with respect to the number of transformation items. (b) The SPICE values with respect to different categories of data.}
\label{fig:spice_diff_setting}
\end{figure}

\begin{figure}
\includegraphics[width=\linewidth]{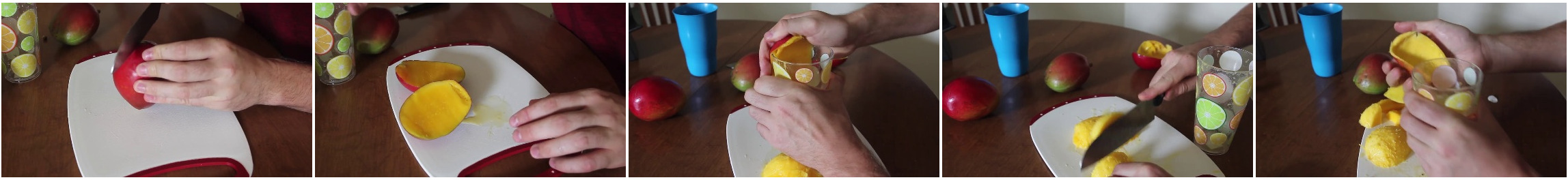}

\begin{tiny}

\centering
\vspace{2pt}

\setlength{\tabcolsep}{5pt}
\begin{tabular}{lrrrr}

\textbf{Groundtruth} & Remove the core.
& Cut the flesh.
& Cut into strips and pieces.
& Cut the flesh. \\

\textbf{Gemini-1.5}  & Cut mango in half.
 & \textcolor{azure(colorwheel)}{Scoop mango with mango skin.}
 & Slice mango flesh.
 & \textcolor{red}{Scoop mango with knife.} \\

\textbf{GPT-4o}  & Place mango on cutting board.
 & \textcolor{orange}{Slice mango with knife.}
 & \textcolor{orange}{Separate mango slices from seed.}
 & \textcolor{orange}{Transfer mango slices to glass.} \\

\textbf{Qwen-VL-chat} & Cutting the fruit.
 & Peeling the fruit.
 & \textcolor{red}{Squeezing the fruit.}
 & \textcolor{red}{Drinking the fruit juice.} \\

\textbf{LLaVA-1.5-7b} & Cut the fruit into pieces.
 & \textcolor{red}{Squeeze the juice into a glass.}
 & \textcolor{orange}{Add a slice of lemon.}
 & \textcolor{orange}{Serve the juice with a straw.} \\

\end{tabular}
\end{tiny}

\vspace{6pt}

\includegraphics[width=\linewidth]{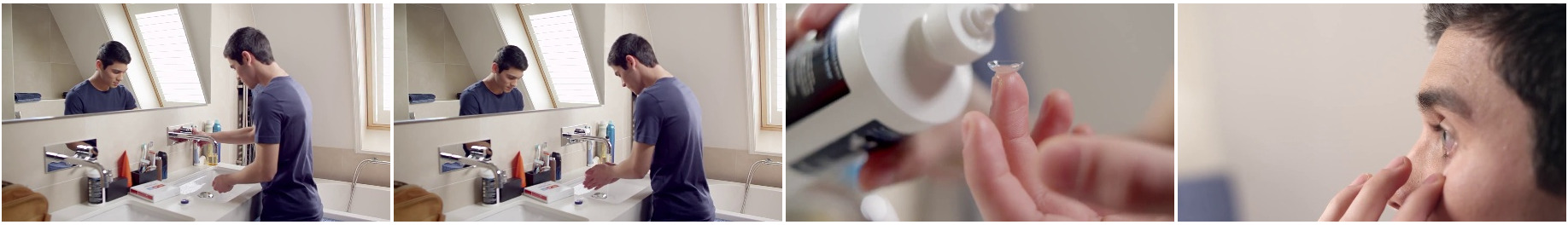}

\begin{tiny}

\centering
\vspace{2pt}

\setlength{\tabcolsep}{8pt}
\begin{tabular}{lrrr}
\textbf{Groundtruth} & Wet and wash hands. & Add some cleaner to clean and wet the lenses and take out the lenses. & Wear the contact lenses. \\

\textbf{Gemini-1.5} & Cup hands under faucet. & Dispense contact solution on finger. & Place contact lens on eye. \\

\textbf{GPT-4o} & \textcolor{orange}{Open contact lens case.} & Pick up contact lens. & Move contact lens to eye. \\

\textbf{Qwen-VL-chat} & \textcolor{orange}{Apply contact lens.} & \textcolor{orange}{Look in mirror.} & \textcolor{orange}{Remove contact lens.} \\

\textbf{LLaVA-1.5-7b} & \textcolor{ao(english)}{Apply lotion to hand.} & \textcolor{ao(english)}{Squeeze bottle of lotion.} & \textcolor{ao(english)}{Spread lotion on hand.}

\end{tabular}
\end{tiny}
\caption{Qualitative comparison on the VTT test data. Above: cut mango. Below: wear contact lenses. Different error types are marked with different colors: bias (\textcolor{red}{red}), misidentification (\textcolor{ao(english)}{green}), hallucination (\textcolor{orange}{orange}), and illogicality (\textcolor{azure(colorwheel)}{blue}).}
\label{fig:error_type}
\end{figure}

\subsection{Comparison of Baseline Models}
Table~\ref{tab:baseline} summarizes the results of models on the VTT dataset, including traditional visual story telling models and MLLMs. 
The results show that both traditional models and SOTA MLLMs have much room for improvement. For traditional models, GLACNet performs best, which chieves $4.75$, $3.82$ and $3.78$ (out of 5) on Fluency, Relevance and Logical Soundnes respectively. This may because GLACNet uses contextual information more completely. For multimodal models, the best performer is Gemini-1.5 which achieves $4.95$, $3.95$ and $4.17$ on Fluency, Relevance and Logical Soundnes respectively. But it does not show a large advantage gap over traditional models. Other MLLMs even perform worse than traditional models, although they have more parameters and use more training data. Further analysis based on human evaluation shows that the main problem with the current large model is inconsistency with the input image, that is, they always generate text that is not completely related or even completely unrelated to the image. In addition, the output of MLLMs also have logical errors, which are manifested in the generated activities violating commonsense or the generated transformations sequence is unreasonable. Even tuning cannot solve these problems well, indicating that more efforts are needed.

We also analyze the performance of the model with different data types. We find that, as illustrated in Figure~\ref{fig:spice_diff_setting}, for all MLLMs, an increase in variation leads to a decrease in model performance. It indicates that the model faces challenges in handling long contexts. We also analyze the performance of the models on different event categories. We find that models are best at different data categories, which may be due to differences in the training data distribution. However, the data category that the models are least good at is the same, i.e., sports. More relevant data may need to be utilized to improve model performance on this type.


\subsection{Qualitative Analysis and Common Error Types}

We qualitatively analyze the output of different MLLMs and show some examples in Figure~\ref{fig:error_type} (more cases can be found at Appendix). We summarize the common errors into four types:

\textbf{Bias:} Models can be misled by the presence of specific objects to conclude that certain non-occurring events are happened. As the example of the event `cut mango', the simultaneous appearance of the glass and the fruit leads the Qwen and LLaVa to assume that the event is related to juicing. This type of error indicates that the models are overly reliant on co-occurrence patterns observed in the training data, which may not accurately reflect real-world scenarios.

\textbf{Misidentification:} Models sometimes mistakenly identify objects in images. For instance, LLava failed to recognize contact lenses and incorrectly identified cleaner as lotion. Such recognition errors are more prevalent in models with smaller parameters. This suggests that model capacity and the training data quality significantly impact the object recognition capability, highlighting the necessity for both larger models and more diverse and comprehensive datasets.

\textbf{Hallucination:} Models sometimes generate predictions that deviate from the image context, despite they correctly identify objects and topics. This results in the generation that is relevant to the topic but inconsistent with the image, or even generating objects that do not exist. As the example of the event `wear contact lenses', the output of GPT-4o is consistent with the topic but includes `contact lens case', which is not present in the image. This issue points to a disconnect between the language and vision components of current MLLMs.

\textbf{Illogicality:} Models may output illogical content or even violate commonsense. For example, Gemini outputs `scoop mangoes with mango skin', which is an implausible scenario. These errors highlight the limitations of models in understanding and applying commonsense reasoning, indicating a need for incorporating more advanced reasoning capabilities and better grounding in real-world knowledge.

\subsection{Further Exploration} 

Inspired by the human cognitive process, we try to improve the existing model to initially explore how to enhance the models' capacity for visual transformation reasoning. Considering flexibility and computational overhead, we improve on the best-performing traditional model GLACNet~\cite{gonzalez-ricoContextualizeShowTell2018,kimGLACNetGLocal2019}. In addition, we replace the image encoder with CLIP~\cite{radfordLearningTransferableVisual2021} to ensure the quality of image understanding. The new model is named TTNet.

The following three areas were investigated: (1) \textbf{Difference sensitive encoding:} besides the original representation of each state, we also treat the differences between every two adjacent states as input to enhance the model's ability to capture semantic-level differences in states. (2) \textbf{Masked transformation modeling (MTM):} to help models fully leverage all the state information and transformations in other steps to reason the current transformation, we adopted a strategy of masked transformation modeling, which is similar to mask language modeling. (3) \textbf{Auxiliary learning:} we add a topic prediction and category prediction task for each state series to enhance the consistency of model outputs and themes. See the Appendix for more details.

\begin{table}[t]
\begin{minipage}[c]{.51\textwidth}
    \centering
    \small
    \setlength{\tabcolsep}{2pt}
      \caption{Results of applying different key components of TTNet. The first row presents the base model's performance.}
    \label{tab:ablation_1}  
    \begin{tabular}{lllrrrrr}
        \toprule
        diff. & MTM & aux. & B@4            & M              & R              & C               & BS             \\
        \midrule
                     &              &          & 55.68          & 60.47          & 67.05          & 515.12          & 72.22          \\
        \midrule
        $\surd$      &              &          & 59.89          & 64.61          & 70.30          & 556.85          & 75.00          \\
                     & $\surd$      &          & 56.26          & 60.92          & 67.57          & 520.04          & 72.72          \\
                     &              & $\surd$  & 56.37          & 61.18          & 67.85          & 521.93          & 72.97          \\
        \midrule
        $\surd$      & $\surd$      &          & 60.39          & 65.38          & 70.99          & 562.25          & 75.62          \\
        $\surd$      &              & $\surd$  & 60.38          & 65.50          & 71.14          & 562.83          & 75.72          \\
                     & $\surd$      & $\surd$  & 56.91          & 61.89          & 68.45          & 527.62          & 73.54          \\
        \midrule
        $\surd$      & $\surd$      & $\surd$  & \textbf{61.22} & \textbf{66.31} & \textbf{71.84} & \textbf{570.63} & \textbf{76.25} \\
        \bottomrule
    \end{tabular}

\end{minipage}
\hspace{3pt}
\begin{minipage}[c]{.46\textwidth}


    \centering
    \small
    \setlength{\tabcolsep}{2pt}
    \caption{Ablation study results on the auxiliary tasks, i.e. category prediction, and topic prediction.}
    \label{tab:ablation_2}
    \begin{tabular}{llrrrrr}
        \toprule
        category    & topic    & B@4            & M              & R              & C               & BS             \\
        \midrule
                &         & 60.39          & 65.38          & 70.99          & 562.25          & 75.62          \\
        \midrule
        $\surd$ &         & 59.11          & 64.08          & 69.99          & 549.44          & 74.81          \\
                & $\surd$ & 60.49          & 65.51          & 71.25          & 562.96          & 75.89          \\
        \midrule
        $\surd$ & $\surd$ & \textbf{61.22} & \textbf{66.31} & \textbf{71.84} & \textbf{570.63} & \textbf{76.25} \\
        \bottomrule
    \end{tabular}

\end{minipage}
\end{table}

Results are shown in Table~\ref{tab:ablation_1}. The outcomes reveal that using the difference feature yields the most significant improvement, implying that the difference is crucial for resolving transformation reasoning. The subsequent four rows demonstrate the results of combining these strategies, and we conclude that employing all three strategies leads to optimal performance. We also examine the effect of different auxiliary tasks. From Table~\ref{tab:ablation_2}, topic classification is more effective than category classification, since topics are more granular than categories. Supervision with two classification tasks simultaneously improves the overall performance, e.g. 562.25 to 570.63 in terms of CIDEr. 

We also try to use improved strategies for LLaVA. Considering both `difference sensitive encoding' and `MTM' require fine-tuning the model parameters to adapt to unseen inputs during pretraining, we only try `auxiliary learning', i.e., also predict the corresponding topic. As in Table~\ref{tab:baseline}, auxiliary learning improves performance whether under zero-shot setting or tuned. Experiment on traditional models shows that explicitly modeling the differences between the states has substantial improvement. But applying similar modeling to MLLMs is not trivial. We leave other improvements to the MLLMs for future work.

\section{Conclusion and Discussion}
\label{sec:conclusion}

This paper introduces visual transformation telling (VTT), a new visual reasoning task that focuses on reasoning transformations between states in a series of images, which is a crucial cognitive skill for humans. To the best of our knowledge, this is the first real-world application for transformation reasoning by defining transformation descriptions as output. We built the VTT dataset using 13,547 samples collected from CrossTask and COIN. We extensively test the capabilities of existing models, both traditional models and state-of-the-art MLLMs. Our experimental results show that even the most advanced large language models cannot solve this task well. We summarise the error types of existing models and find that current errors are mainly concentrated in four aspects, i.e., bias, misidentification, hallucination and illogicality. In addition, we conduct extensive experiments by tuning MLLMs on VTT data, prompting to force topic generation, and proposing several strategies for traditional models. 
We believe that collecting and providing more data with transformation information and adapting MLLMs to understand differences between states(images) are the most promising research directions for future study of transformation reasoning.


\paragraph{Limitation.} Our VTT dataset covers a limited range of transformations, which limits the models' applicability of visual transformation reasoning. Additionally, the limited size of the VTT dataset hinders the generalization ability of current models. However, collecting a larger dataset is costly due to the expense of annotating steps/transformations with descriptions and temporal boundaries. One possible way to mitigate this cost is to use pretrained step localization models~\cite{wangSelfSupervisedLearningSemiSupervised2021, zhangActionFormerLocalizingMoments2022} or action and object state-recognition models~\cite{soucekLookChangeLearning2022a} to propose coarse steps/transformations and further refine the results with human annotators. In addition, we suggest using object state-recognition tools~\cite{soucekLookChangeLearning2022a} to refine the boundary precision of existing step segments in CrossTask and COIN for constructing larger datasets in future studies. 

\textbf{Broader impact.} VTT can support many real-world applications that require the understanding of visual transformation reasoning. For example, in video generation, a better understanding of the transformation between states (key frames) helps to generate more coherent and reasonable videos. Furthermore, VTT also helps models to master embodied intelligence better, such as procedural planning or intelligent robots. In order to interact with the world and perform specific tasks, it is necessary to fully understand what transformations are required to achieve a desired state. 

\textbf{Ethical considerations.} The images in our dataset contain people's activities and personal information such as facial information. However, considering that all data is collected from public datasets, there is no violation of privacy.




\bibliographystyle{unsrt}

\providecommand{\noopsort}[1]{}

\newpage

\section*{Checklist}


\begin{enumerate}

\item For all authors...
\begin{enumerate}
  \item Do the main claims made in the abstract and introduction accurately reflect the paper's contributions and scope?
    \answerYes{See Section~\ref{sec:introduction}.}
  \item Did you describe the limitations of your work?
    \answerYes{See Section~\ref{sec:conclusion}.}
  \item Did you discuss any potential negative societal impacts of your work?
    \answerYes{See Section~\ref{sec:conclusion}.}
  \item Have you read the ethics review guidelines and ensured that your paper conforms to them?
    \answerYes{}
\end{enumerate}

\item If you are including theoretical results...
\begin{enumerate}
  \item Did you state the full set of assumptions of all theoretical results?
    \answerNA{}
	\item Did you include complete proofs of all theoretical results?
    \answerNA{}
\end{enumerate}

\item If you ran experiments (e.g. for benchmarks)...
\begin{enumerate}
  \item Did you include the code, data, and instructions needed to reproduce the main experimental results (either in the supplemental material or as a URL)?
    \answerYes{See Section~\ref{sec:introduction} and supplemental material.}
  \item Did you specify all the training details (e.g., data splits, hyperparameters, how they were chosen)?
    \answerYes{See supplemental material.}
	\item Did you report error bars (e.g., with respect to the random seed after running experiments multiple times)?
    \answerNo{Experimental results are stable.}
	\item Did you include the total amount of compute and the type of resources used (e.g., type of GPUs, internal cluster, or cloud provider)?
    \answerYes{See supplemental material.}
\end{enumerate}

\item If you are using existing assets (e.g., code, data, models) or curating/releasing new assets...
\begin{enumerate}
  \item If your work uses existing assets, did you cite the creators?
    \answerYes{See Section~\ref{sec:model_selection} and supplemental material.}
  \item Did you mention the license of the assets?
    \answerYes{See supplemental material.}
  \item Did you include any new assets either in the supplemental material or as a URL?
    \answerYes{See supplemental material.}
  \item Did you discuss whether and how consent was obtained from people whose data you're using/curating?
    \answerYes{See supplemental material.}
  \item Did you discuss whether the data you are using/curating contains personally identifiable information or offensive content?
    \answerYes{See Section~\ref{dataset_construction}}
\end{enumerate}

\item If you used crowdsourcing or conducted research with human subjects...
\begin{enumerate}
  \item Did you include the full text of instructions given to participants and screenshots, if applicable?
    \answerYes{See supplemental material.}
  \item Did you describe any potential participant risks, with links to Institutional Review Board (IRB) approvals, if applicable?
    \answerNA{}
  \item Did you include the estimated hourly wage paid to participants and the total amount spent on participant compensation?
    \answerYes{See supplemental material.}
\end{enumerate}

\end{enumerate}

\newpage

\appendix

\section{Dataset Scale Discussion}

\begin{table*}[t]
\centering

\scriptsize
\setlength{\tabcolsep}{4pt}
\begin{tabular}{p{0.15\linewidth} p{0.8\linewidth}}
\toprule
Category & Topics \\
\midrule

Nursing and Care (14) & Wash Dog, Use Earplugs, Use Neti Pot, Put On Hair Extensions, Use Epinephrine Auto-injector, Perform CPR, Wear Contact Lenses, Remove Blackheads With Glue, Give An Intramuscular Injection, Shave Beard, Wash Hair, Bandage Dog Paw, Draw Blood, Bandage Head \\
\midrule
Pets and Fruit (7) & Plant Tree, Transplant, Graft, Cut Grape Fruit, Cut Mango, Cut Cantaloupe, Sow \\
\midrule
Furniture and Decoration (15) & Install Shower Head, Install Ceramic Tile, Install Air Conditioner, Install Curtain, Lubricate A Lock, Replace Door Knob, Install Wood Flooring, Install Closestool, Assemble Cabinet, Assemble Sofa, Replace Faucet, Replace Toilet Seat, Assemble Bed, Build Simple Floating Shelves*, Assemble Office Chair \\
\midrule
Leisure and Performance (17) & Make Paper Wind Mill, Perform Vanishing Glass Trick, Raise Flag, Play Frisbee With A Dog, Make Chinese Lantern, Carve Pumpkin, Change Guitar Strings, Perform Paper To Money Trick, Pitch A Tent, Open Champagne Bottle, Blow Sugar, Make Paper Easter Baskets, Cut And Restore Rope Trick, Do Lino Printing, Replace Drumhead, Prepare Sumi Ink, Prepare Canvas \\
\midrule
Dish (23) & Make Kimchi Fried Rice*, Cook Omelet, Make Sandwich, Grill Steak*, Clean Fish, Use Toaster, Clean Shrimp, Make Burger, Make French Toast*, Wrap Zongzi, Make French Strawberry Cake*, Make Pickles, Boil Noodles, Make Bread and Butter Pickles*, Make Kerala Fish Curry*, Make Lamb Kebab, Make French Fries, Use Rice Cooker To Cook Rice, Make Pizza, Make Youtiao, Make Salmon, Smash Garlic, Make Pancakes* \\
\midrule
Electrical Appliance (20) & Replace Graphics Card, Replace Light Socket, Replace Electrical Outlet, Replace Memory Chip, Use Soy Milk Maker, Change Toner Cartridge, Replace Laptop Screen, Replace Refrigerator Water Filter, Use Vending Machine, Replace Filter For Air Purifier, Replace Hard Disk, Replace Blade Of A Saw, Refill Cartridge, Clean Laptop Keyboard, Arc Weld, Install Ceiling Fan, Replace A Bulb, Paste Screen Protector On Pad, Assemble Desktop PC, Use Sewing Machine \\
\midrule
Science and Craft (15) & Prepare Standard Solution, Make Flower Press, Use Volumetric Pipette, Hang Wallpaper, Make Candle, Make Soap, Use Triple Beam Balance, Make Flower Crown, Use Volumetric Flask, Paste Car Sticker, Make Slime With Glue, Make Paper Dice, Wrap Gift Box, Set Up A Hamster Cage, Use Analytical Balance \\
\midrule
Drink and Snack (20) & Make Meringue*, Make Salad, Make Lemonade*, Make Taco Salad*, Make Tea, Make Chocolate, Make a Latte*, Make Homemade Ice Cream, Make Jello Shots*, Make Coffee, Make Cocktail, Make Cookie, Make Irish Coffee*, Roast Chestnut, Make Banana Ice Cream*, Make Orange Juice, Make Matcha Tea, Make Sugar Coated Haws, Make Strawberry Smoothie, Make Hummus \\
\midrule
Vehicle (21) & Change Bike Chain, Replace Car Fuse, Replace Rearview Mirror Glass, Tie Boat To Dock, Pump Up Bicycle Tire, Change Car Tire, Use Jack, Remove Scratches From Windshield, Jack Up a Car*, Change Bike Tires, Install License Plate Frame, Fuel Car, Replace A Wiper Head, Install Bicycle Rack, Replace Tyre Valve Stem, Change a Tire*, Patch Bike Inner Tube, Polish Car, Replace Car Window, Add Oil to Your Car*, Park Parallel \\
\midrule
Housework (15) & Put On Quilt Cover, Clean Bathtub, Wash Dish, Clean Leather Seat, Pack Sleeping Bag, Clean Wooden Floor, Clean Toilet, Iron Clothes, Drill Hole, Remove Crayon From Walls, Clean Hamster Cage, Make Bed, Unclog Sink With Baking Soda, Clean Rusty Pot, Clean Cement Floor \\
\midrule
Sport (10) & Practise Karate, Wear Shin Guards, Practise Triple Jump, Throw Hammer, Play Curling, Practise Skiing Aerials, Practise Pole Vault, Attend N B A Skills Challenge, Glue Ping Pong Rubber, Practise Weight Lift \\
\midrule
Gadgets (21) & Open A Lock With Paperclips, Replace Mobile Screen Protector, Load Grease Gun, Change Mobile Phone Battery, Replace Sewing Machine Needle, Change Battery Of Watch, Replace SIM Card, Resize Watch Band, Replace CD Drive With SSD, Refill Mechanical Pencils, Make Wireless Earbuds, Refill Fountain Pen, Refill A Lighter, Rewrap Battery, Replace Battery On Key To Car, Fix Laptop Screen Scratches, Operate Fire Extinguisher, Replace Battery On TV Control, Use Tapping Gun, Refill A Stapler, Make RJ45 Cable \\

\bottomrule
\end{tabular}
\caption{The Categories and topics in VTT dataset. Topics marked with * are from CrossTask and others belong to COIN.}
\label{tab:topic_list}
\end{table*}

As mentioned in the main paper, the limited size of the VTT dataset hinders the generalization ability of current models. Additionally, the dataset covers only a narrow range of transformations, which limits the models' applicability. However, collecting a larger dataset is costly due to the expense of annotating steps/transformations with descriptions and temporal boundaries are expensive. One possible way to mitigate this cost is to use pretrained step localization models~\cite{wangSelfSupervisedLearningSemiSupervised2021, zhangActionFormerLocalizingMoments2022} or action and object state-recognition models~\cite{soucekLookChangeLearning2022a} to propose coarse steps/transformations and refine the results with human annotators. In addition, we suggest using object state-recognition~\cite{soucekLookChangeLearning2022a} to refine the boundary precision of existing step segments in CrossTask and COIN for constructing larger datasets in the future. Apart from annotating a large-scale dataset, another way is to design a method that can directly learn transformation reasoning from massive raw video-caption data such as HowTo100M~\cite{miechHowTo100MLearningTextVideo2019}. There have already been pioneer works that obtain impressive results on natural language processing tasks, such as GPT-3~\cite{brownLanguageModelsare2020} and chatGPT~\footnote{\url{https://chat.openai.com/}}, and computer tasks, such as CLIP~\cite{radfordLearningTransferableVisual2021}.

\section{The Categories and Topics in VTT }

Each sample in VTT has a topic and a category. All Categories and topics are shown in Table~\ref{tab:topic_list}.

\begin{table}[t]
    \centering
    \scriptsize
    \setlength{\tabcolsep}{3pt}
    \begin{tabular}{c c p{0.7\linewidth}}
    \toprule
         Metric & Score & Criteria  \\
    \midrule
          \multirow[c]{6}{*}{\rotatebox[origin=c]{90}{\makecell{Fluency}}}
          & 5 & All sentences are fluent. \\
          & 4 & Most sentences are fluent, with only a few flaws. \\
          & 3 & About half of the sentences are fluent. \\
          & 2 & Most of the sentences are difficult to read, with only a few being okay. \\
         & 1 & All sentences are hard to read. \\ 
    \midrule
         \multirow[c]{12}{*}{\rotatebox[origin=c]{90}{\makecell{Relevance}}}
          & 5 & The descriptions are all related to the corresponding before and after images.  \\
          & 4 & A few descriptions are slightly irrelevant, e.g. the description is related to the underlying topic but cannot be clearly inferred from the images. \\
          & 3 & Many descriptions are slightly irrelevant or a few descriptions are irrelevant, e.g. the action or target object mentioned in the transformation does not match the images.  \\
          & 2 & Many descriptions are irrelevant. \\
          & 1 & Most descriptions are irrelevant, or some descriptions are completely irrelevant, e.g. transformation is unrelated to the underlying topic of the images. \\
    \midrule
         \multirow[c]{11}{*}{\rotatebox[origin=c]{90}{\makecell{Logical\\Soundness}}}
          & 5 & The underlying logic of the descriptions is consistent with common sense. \\
          & 4 & The overall logic is consistent with common sense, with minor flaws. \\
          & 3 & There are a few obvious logical problems between the descriptions, e.g. unresonable repeating transformations. \\ 
          & 2 & There are some obvious logical problems, e.g. the order of transformations is obviously not in line with common sense. \\
          & 1 & Logic cannot be judged because of the extremely poor fluency or poor relevance leading to overall logic inconsistent with the underlying topic. \\
    \bottomrule
    \end{tabular}
        \caption{The VTT human evaluation guidelines.}
    \label{tab:human_evaluation}
\end{table}

\begin{figure*}[t]
    \centering
    \includegraphics[width=0.8\linewidth]{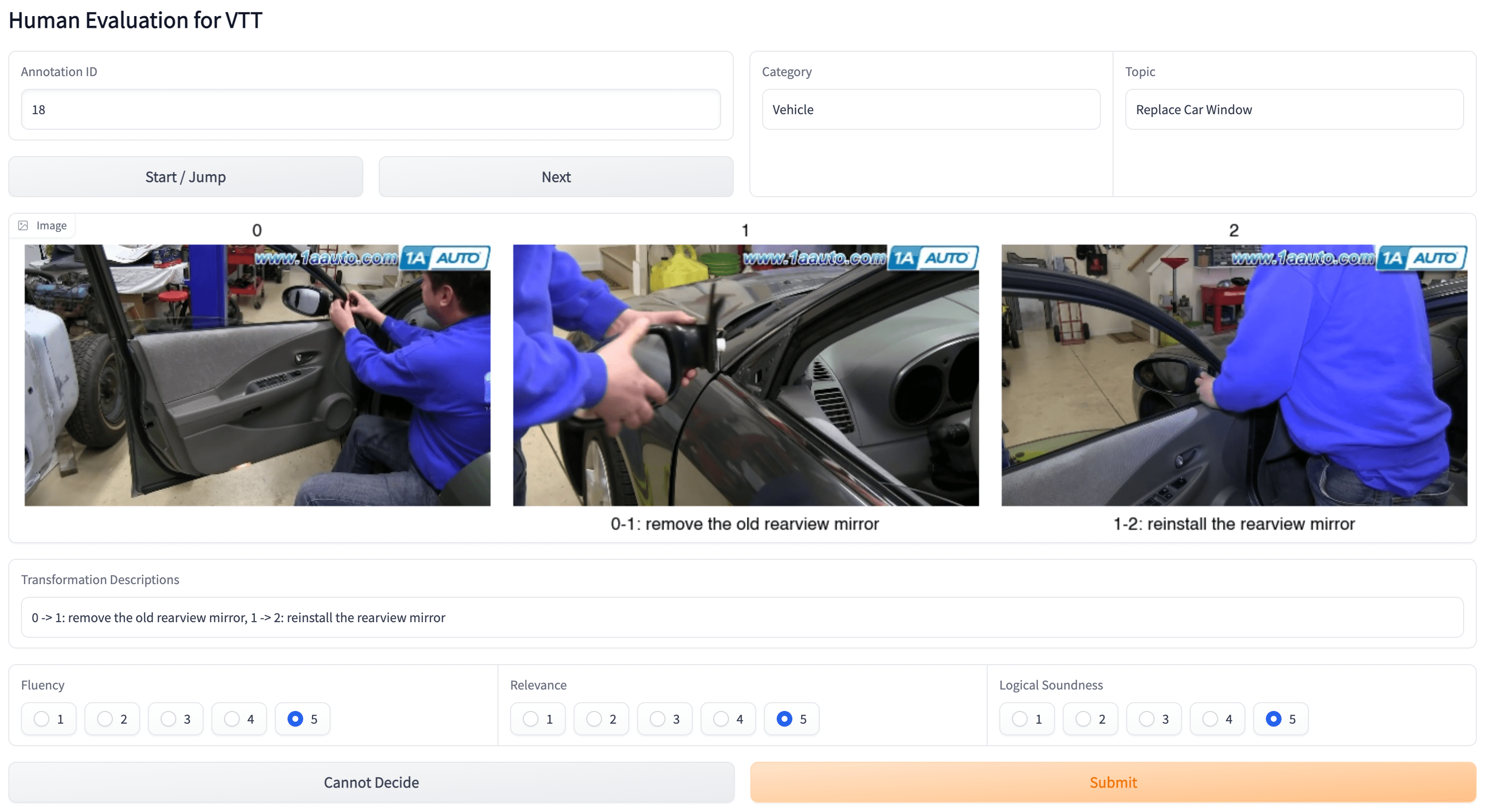}
    \caption{The web interface of human evaluation on VTT.}
    \label{fig:human_evaluation}
\end{figure*}

\section{Evaluation for VTT}


\subsection{Automatic Evaluation}

The computation of BLEU@4 follows the smooth strategy \cite{chenSystematicComparisonSmoothing2014} to improve the accuracy of the results. This is necessary because the descriptions in the VTT dataset are typically short, resulting in a zero score when using the original BLEU@4 method. In addition, BERT-Score is rescaled with the pre-computed baseline~\cite{zhangBERTScoreEvaluatingText2020} to provide more meaningful scores with a wider range. The NLTK package~\footnote{\url{https://www.nltk.org/api/nltk.translate.bleu_score.html}} is used to compute BLEU@4, while CIDEr, METEOR, ROUGE, and SPICE are computed using the code from coco-caption~\footnote{\url{https://github.com/tylin/coco-caption}}. BERT-Score is computed using the official code~\footnote{\url{https://github.com/Tiiiger/bert_score}} provided by the authors.

\subsection{Human Evaluation}
\label{sec:human_eval}
Automatic evaluation metrics have limitations in reflecting the quality of the generated text, as they are uninterpretable and do not necessarily align with human evaluations~\cite{vanderleeBestpracticeshuman2019}. To address this, we manually evaluate text quality in the VTT task using three levels of assessment. The first level assesses the fluency of the text, while the second level evaluates the relevance of each transformation description to the topic and to the images before and after. The third level assesses the logical consistency between transformation descriptions. The assessment is conducted using a 5-point Likert scale and follows the guidelines presented in Table~\ref{tab:human_evaluation}. We invited 25 volunteers to evaluate major baseline models on a subset of 200 samples randomly sampled from the testing set, including one sample from each topic and two additional samples. Annotators were asked to read and follow the guidelines to assign scores. During the human evaluation process, annotators were able to view the images, the category, and the topic as references. At least two individuals evaluated each model's result for each sample. The web interface for human evaluation is shown in Figure~\ref{fig:human_evaluation} and will be included in the VTT source code.

\section{TTNet}
\label{app:ttnet}

Our TTNet is inspired by human's cognitive process of transformation and existing visual storytelling models~\cite{gonzalez-ricoContextualizeShowTell2018,kimGLACNetGLocal2019}. In this section, we first introduce the problem formulation and the basic structure of TTNet. Then we describe how we model transformation by enhancing the model's ability to capture semantic-level differences with difference sensitive encoding, and fully utilize context to strengthen transformation reasoning with masked transformation model and auxiliary learning.

\textbf{Base structure of TTNet.} Inspired by humans and existing visual storytelling models, the first step in TTNet is independent recognition, where each image is understood independently. To achieve this, an \textbf{image encoder} $f_\text{state}$ is introduced to \textit{semantize} each image into a vector, resulting in a set of state representations $V = \{v_i\}_{i=1}^{N+1} = \{f_\text{state}(s_i) \}_{i=1}^{N+1}$. The next step is to associate these states together to form a complete understanding of the event. To reflect this process, a \textbf{context encoder} is used. This encoder, which can be a bi-directional RNN or a transformer encoder, is denoted as $f_\text{trans}$ and \textit{contextualizes} the state representations to obtain transformation representations $C = \{c_i\}_{i=1}^{N+1} = \{f_\text{trans}(i, V) \}_{i=1}^{N+1}$. The final step is to describe the transformations based on the existing understanding. In TTNet, this is achieved using a \textbf{transformation decoder} $f_\text{text}$, which can be an RNN or a transformer decoder. This decoder \textit{textualizes} $N$ transformation representations into separate descriptions $T = \{t_i\}_{i=1}^{N} = \{f_\text{text}(c_{i+1})\}_{i=1}^{N}$, in an auto-regressive manner. Empirically, it was found that adding the transformation representation to the word embedding in each step is better than using it as the prefix token. The training objective is to reduce the gap between generated transformations and ground truth transformations $T^*=\{t^*_i\}_{i=1}^{N}$ by minimizing the negative log-likelihood loss, where $t^*_i=\{x^*_{i,l}\}_{l=1}^{L}$ is the ground truth description of the $i_\text{th}$ transformation.
\begin{equation}
    \mathcal{L}_\text{text} = - \sum_{i=1}^{N} \sum_{l=1}^{L} \log p_{\theta}(x^*_{i,l}|x^*_{i,<l})
\end{equation}

Next, we introduce three strategies we used to model transformation, and we called the model that does not use these strategies as TTNet$_\text{base}$.

\begin{figure}[t]
\centering
\includegraphics[width=0.85\linewidth]{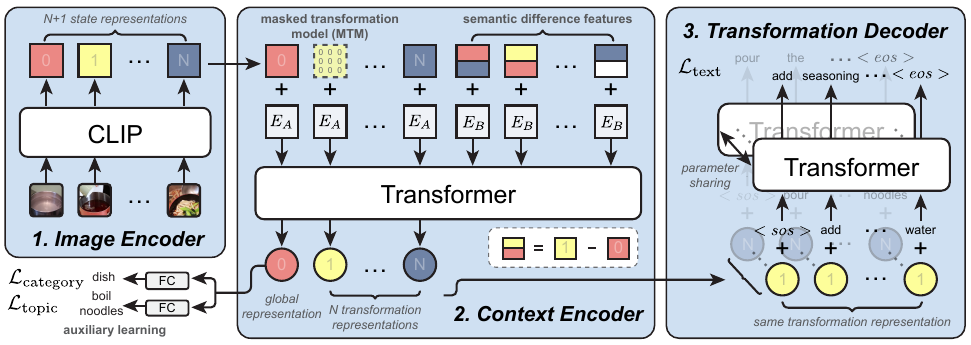}
\caption{\textbf{The architecture of TTNet.} Images are first \textit{semantized} into state representations in the image encoder, then \textit{contextualized} to be transformation representations in the context encoder, and finally \textit{textualized} into text by the transformation decoder. To better modeling transformation, difference sensitive encoding is used to capture semantic-level differences, masked transformation model and auxiliary learning are used to fully utilize context to strengthen transformation reasoning.
}
\label{fig:ttnet}
\end{figure}

\textbf{Difference Sensitive Encoding.} To bridge the semantic gap between state differences and transformation descriptions, the first step is to enable the model to accurately identify and capture the variations between states. However, capturing differences is challenging since adjacent states often exhibit minimal variation at the pixel level. This is mainly because the scene remains almost unchanged before and after the transformation, and only certain attributes of the transformed object have changed. Our intuition to solve this problem is that despite the minimal differences between states at the pixel level, there are often significant semantic differences. Therefore, we first choose CLIP~\cite{radfordLearningTransferableVisual2021} as our image encoder to extract state representations, due to CLIP's strong semantic representation ability trained on large-scale unsupervised data. Then, we compute semantic difference features between adjacent states by subtracting the current state and the previous state representations $\Delta V = \{v_i - v_{i-1}\}_{i=1}^{N+1}$, where $v_0 = v_{N+1}$. In TTNet, we feed both state representations and the semantic difference features into the context decoder. To make the model able to distinguish these two kinds of features, we initialize two learnable types of embeddings and add them to the corresponding features.



\begin{figure}
\small
\includegraphics[width=\linewidth]{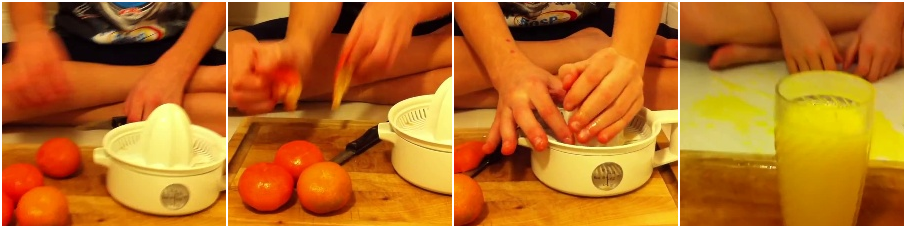}\\
\quad 1. Cut both ends and remove fruit seeds.\\
\quad \textcolor{red}{2. Pour the egg into the bowl.}\\
\quad  3. Pour the orange juice into the cup.

\caption{A failure case from TTNet$_\text{base}$ which has the potential to be corrected by utilizing context information.}
\label{fig:crossmodal_without_context}

\end{figure}

\textbf{Masked Transformation Model.} After identifying state differences, the next challenge is to efficiently reason about the underlying transformations. For humans, one common approach is to fully utilize the context to aid reasoning rather than focusing solely on adjacent states.
Therefore, we chose the transformer~\cite{vaswaniAttentionAllyou2017} as the backbone of the context encoder, given its well-known ability to encode contextual information. However, in our initial experiments, we found TTNet$_\text{base}$ failed to fully utilize context information when reasoning about transformations. A typical example is shown in Figure~\ref{fig:crossmodal_without_context}, where TTNet$_\text{base}$ mistakenly identified an orange as an egg due to their similarities in the image. Nevertheless, such ambiguity can be resolved by incorporating other correct transformations. Hence, the question becomes how to enhance the model's ability to leverage contextual information. Inspired by BERT objectives, we proposed two strategies, including the masked transformation model (MTM) and auxiliary learning. Similar to the masked language model~\cite{devlinBERTPretrainingDeep2019}, the intuition behind MTM is that one transformation can be reasoned from nearby transformations. Specifically, during training, 15\% of the features fed into the context encoder, including state representations and semantic difference features, are randomly masked. Empirically, we found using MTM with a 50\% probability works better.

\textbf{Auxiliary Learning.} Following the target of fully utilizing context information, another strategy is focused on the global representation. BERT applied the objective of next sentence prediction (NSP) but this is not suitable for our task. However, we found humans usually try to guess the category or topic before describing transformations, e.g.~cooking noodles. Therefore, we set another objective that requires TTNet to predict the category and topic from the global representation during training. Two additional cross-entropy losses $\mathcal{L}_\text{category}$ and $\mathcal{L}_\text{topic}$ can be computed from these two classification problems. The final training loss becomes a combination of $\mathcal{L}_\text{text}$, $\mathcal{L}_\text{category}$, and $\mathcal{L}_\text{topic}$, with adjustment factor $\alpha$ and $\beta$:
\begin{equation}
    \mathcal{L} = \mathcal{L}_\text{text} + \alpha \mathcal{L}_\text{category} + \beta \mathcal{L}_\text{topic}.
\end{equation}

\begin{table*}[t]
\centering
\caption{Implementations details of baseline models and TTNet.}
\label{tab:model_implementation}
\small

\begin{tabular}{lrrrr}
\toprule
Model & Image Encoder & Context Encoder & Transformation Decoder & Params  \\
\midrule
CST & InceptionV3 & LSTM & LSTM  & 379M \\
CST* & CLIP (ViT-L/14)  & LSTM & LSTM & 661M  \\
GLACNet & ResNet152 & bi-LSTM &  LSTM & 128M \\
GLACNet* & CLIP (ViT-L/14) & bi-LSTM & LSTM & 373M \\
DenseCap* & CLIP (ViT-L/14) & Attention & LSTM & 361M \\
\midrule
TTNet$_\text{Base}$ & CLIP (ViT-L/14) & Transformer & Transformer & 368M \\
TTNet & CLIP (ViT-L/14) & Transformer & Transformer & 368M \\
\bottomrule
\end{tabular}
\end{table*}

\begin{table}[t]
\centering
\small
\captionof{table}{Results of different image encoders.}
\label{tab:image_encoder}
\begin{tabular}{llrr|rrr}
\toprule
\multicolumn{2}{c}{Image Encoder}& Params & Acc & B@4 & C & BS \\
\midrule
\multirow[c]{5}{*}{\rotatebox[origin=c]{90}{\makecell{ImageNet\\Pretrained\footnotemark}}} & InceptionV3~\cite{szegedyRethinkingInceptionArchitecture2016} & 23M & 77.44 & 44.88 & 404.85 & 61.75 \\
 & ResNet152~\cite{heDeepResidualLearning2016} & 59M & 82.82 & 50.71 & 464.01 & 67.40 \\
 & ViT-L~\cite{dosovitskiyImageWorth16x162022} & 304M & 85.84 & 58.26 & 540.46 & 73.59 \\
 & Swin-L~\cite{liuSwinTransformerHierarchical2021} & 196M & 86.32 & 57.36 & 531.51 & 73.03 \\
 & BEiT-L~\cite{baoBEiTBERTPreTraining2022} & 306M & 87.48 & 41.57 & 370.00 & 58.80 \\
\midrule
\multirow[c]{5}{*}{\rotatebox[origin=c]{90}{\makecell{Image-text\\Pretrained\footnotemark}}} & RN50 & 39M & 73.30 & 53.35 & 491.80 & 69.79 \\
 & RN101 & 57M & 75.70 & 53.78 & 495.30 & 70.08 \\
 & ViT-B/32 & 88M & 76.10 & 55.21 & 510.08 & 71.27 \\
 & ViT-B/16 & 86M & 80.20 & 57.73 & 534.92 & 73.37 \\
 & ViT-L/14 & 304M & 83.90 & \textbf{61.22} & \textbf{570.63} & \textbf{76.25} \\
\bottomrule
\end{tabular}
\end{table}

\footnotetext[7]{Model weights and top-1 accuracy on ImageNet of ImageNet pretrained models are from: \url{https://github.com/rwightman/pytorch-image-models}}
\footnotetext[8]{Pretrained weights of CLIP models are from \url{https://github.com/openai/CLIP} and top-1 accuracy on ImageNet is from Table 10 of the original paper.}

\section{Implementation Detail of Models} \label{sec:implementation}

\subsection{Traditional Models}

The training process of includes standard image augmentation techniques such as random cropping and flipping, resulting in images cropped into 224$\times$224 patches. The architectures of all baseline models are presented in Table~\ref{tab:model_implementation}.

We re-implemented CST and GLACNet based on the original papers and their released source code~\footnote{\url{https://github.com/dianaglzrico/neural-visual-storyteller}}~\footnote{\url{https://github.com/tkim-snu/GLACNet}}. We followed the paper for implementing the final model of DenseCap since we could not find its code. However, we used CLIP to replace DenseCap's original video encoder because it was designed for video descriptions.

The implementation of TTNet includes a default CLIP image encoder of ViT-L/14, which is pretrained and fixed during training. We compare multiple other image encoders in Section~\ref{sec:image_encoder}. The context encoder uses a transformer-based architecture consisting of two transformer encoder layers, implemented using x-transfomer~\footnote{\url{https://github.com/lucidrains/x-transformers}}. All transformer layers use simplified relative positional encoding~\cite{raffelExploringLimitsTransfer2020}. In the transformation decoder part, we directly borrow CLIP's tokenizer and their vocabulary list. Each transformation description is generated separately with a shared two-layer transformer decoder. The idea of adding transformation representations into word embeddings is inspired by GLACNet~\cite{kimGLACNetGLocal2019} and we empirically found this way improves a lot on language influence compared with using the representation as the start token. Like the context encoder, simplified relative positional encoding is also used in the transformation decoder.

We use top-$k$ top-$p$ sampling with $k=100$ and $p=0.9$ to generate text. The dimension of intermediate vectors, including state representations, transformation representations, and word embeddings, is set to 512.
For the training loss, we set the adjustment factor $\alpha$ for $\mathcal{L}_\text{category}$ to 0.025 and $\beta$ for $\mathcal{L}_\text{topic}$ to 0.1. We use the AdamW optimizer~\cite{loshchilovDecoupledWeightDecay2022}, with a learning rate that warms up to 1e-4 in the first 2000 steps and then gradually decreases to 0. All models are implemented with PyTorch~\cite{paszkePyTorchImperativeStyle2019} and trained on a single Tesla A100 80G GPU card with 50 epochs. The code will be released publicly.

\subsection{Multimodal Language Models}

\begin{figure}
\small
\includegraphics[width=\linewidth]{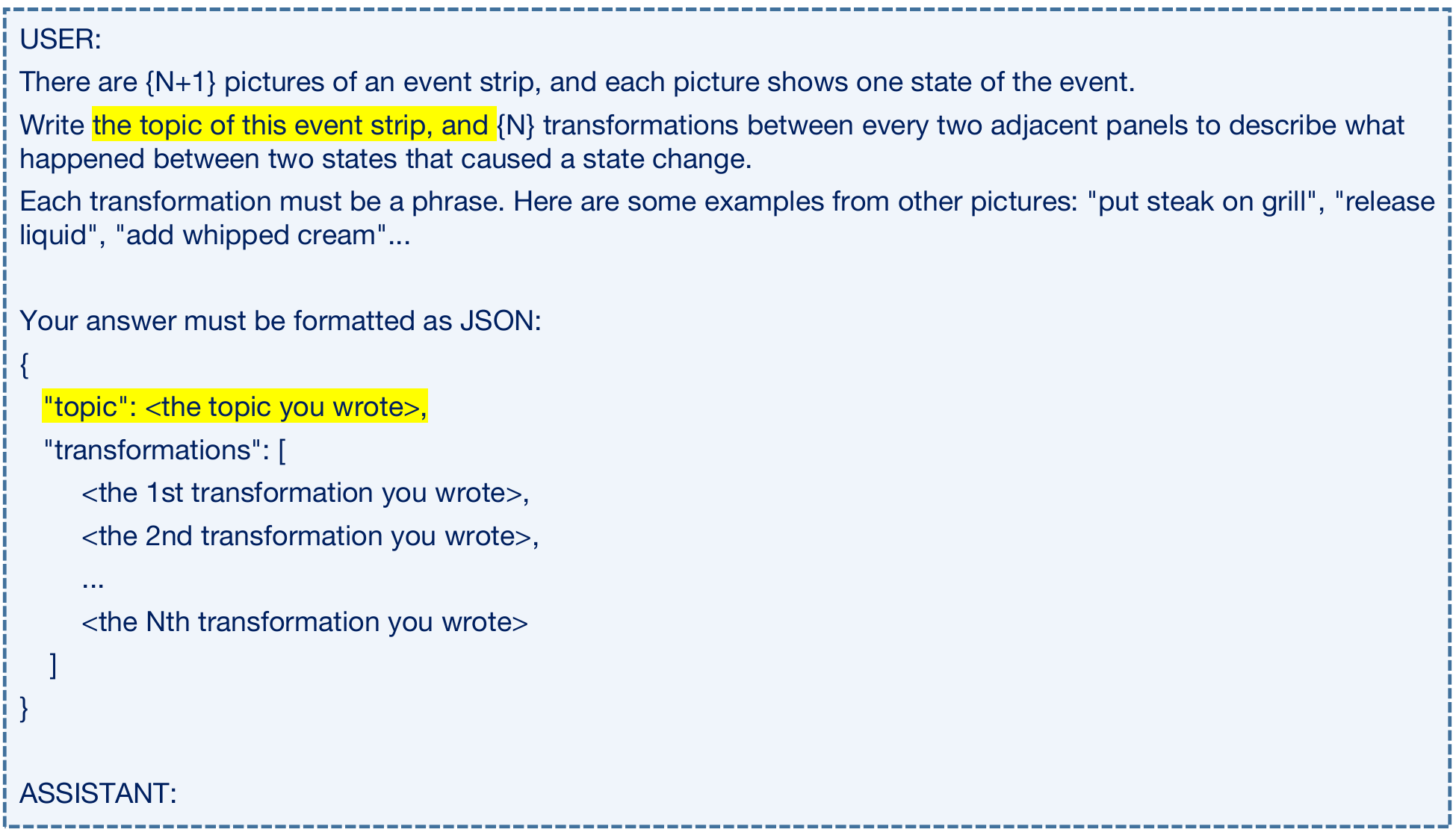}
\caption{Template used to generate prompts for testing multimodal language models. The content highlighted in yellow is only used when adding a topic prediction task, it is not included in the prompt in the standard setting. }
\label{fig:prompt}
\end{figure}

To establish MLLMs performance and provide fair comparisons, we employ the exact same
prompting structure as in Figure~\ref{fig:prompt}, in which $N$ should be replaced to the transformation number. Since existing pretrained MLLMs (except Qwen) either do not support multiple image inputs or perform poorly when processing multiple images in order, we adapted the model's input requirements by collapsing the multiple images corresponding to each sample into a single one.
We follow the official implementation~\footnote{\url{https://github.com/haotian-liu/LLaVA/blob/main/scripts/v1_5/finetune_lora.sh}} to tune LLaVA with LORA. We conduct our experiments over 50 epochs, employing a batch size of 16. The learning rate is set to 2e-5 and the warmup ratio is 0.03.

\section{More Analyses on TTNet}


\begin{table}[t]
    \centering

    \caption{Results of different strategies of computing difference features.}
    \label{tab:crossmodal_diff}
    \small
    \begin{tabular}{llrrrrr}
        \toprule
        state & diff & B@4            & M              & R              & C               & BS             \\
        \midrule
        $\surd$  & -            & 56.91          & 61.89          & 68.45          & 527.62          & 73.54          \\
        \midrule
        $\surd$  & early     & 60.10          & 65.16          & 70.88          & 559.78          & 75.69          \\
        $\surd$  & late     & \textbf{61.22} & \textbf{66.31} & \textbf{71.84} & \textbf{570.63} & \textbf{76.25} \\
        \bottomrule
    \end{tabular}
\end{table}

\subsection{Comparison of Early and Later Differences}
In the main paper, we computed the difference features in a later fusion manner, i.e., computing them on encoded image vectors to produce the semantic difference. In this section, we compare this approach with an the alternative one, early fusion, which calculates pixel-level difference on raw images before feeding them to the image encoder. In TVR~\cite{hongTransformationDrivenVisual2021}, early differences were found to be more effective, while Table~\ref{tab:crossmodal_diff} shows the opposite result. We explain that this is because TVR involves predicting property changes on synthetic data, which relies more on pixel differences. In contrast, VTT requires event-level descriptions, placing greater emphasis on semantic distinctions.

\begin{table}[t]
\centering
\caption{Models perform worse with only adjacent states in terms of CIDEr score and re-training on them still falls short of the normal setting.}
\label{tab:crossmodal_independent}
\small
\begin{tabular}{lrr}
\toprule
Model & Normal & Adjacent States Only \\
\midrule
CST* & 84.90 & 49.80 \\
DenseCap* & 439.53 & 295.75 \\
GLACNet* & 508.19 & 268.49 \\
TTNet & \textbf{570.63} & 349.96 \\
\midrule
TTNet (retrain) & - & \textbf{459.84} \\
\bottomrule
\end{tabular}
\end{table}

\subsection{Analyses on Context Modeling} \label{sec:context}


\textbf{Analyzing Context Importance for VTT.} To determine the importance of the context for VTT, we evaluated models in an independent setting where each transformation could only be reasoned from two adjacent states, without accessing other states. If context were not important, the performance of models would remain unchanged. However, Table~\ref{tab:crossmodal_independent} shows all four models experienced a significant performance drop. For example, TTNet's CIDEr score decreased by approximately 39\%, indicating the crucial role of context in transformation reasoning. We also retrained TTNet on data constructed following the independent setting, and while performance improved, there remained a considerable gap compared to fully accessing context, further demonstrating the importance of context for VTT.


\begin{figure}[t]
\centering
\includegraphics[width=0.5\linewidth]{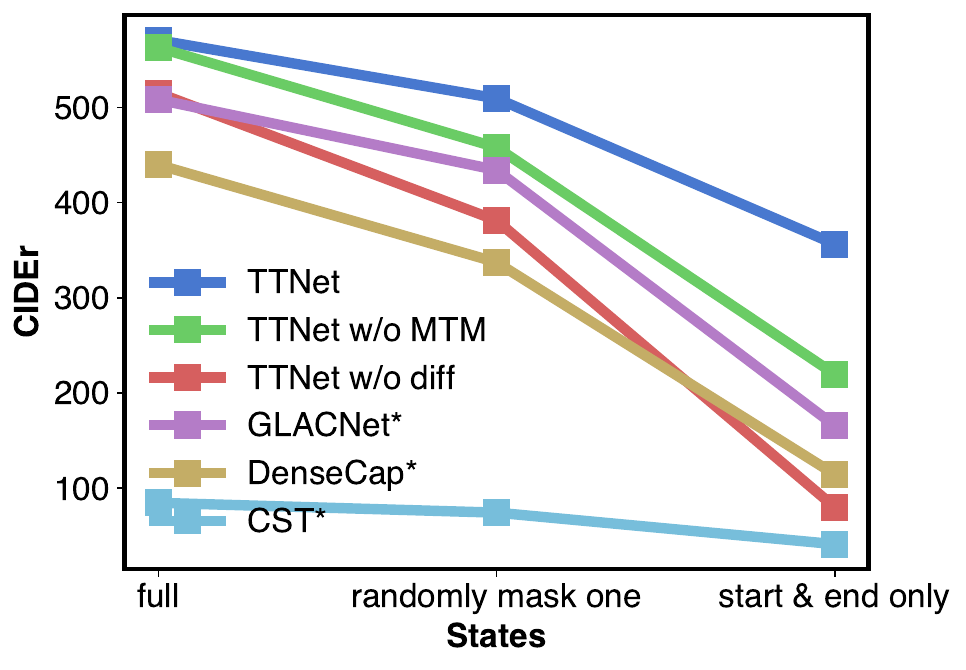}
\caption{TTNet performs most robustly when reasoning on partial context (some states are missing).}
\label{fig:crossmodal_miss}
\end{figure}

\textbf{Assessment on Utilizing Context.} Having established the importance of context, it is important to test models' ability to utilize it. We examined two settings where the provided states gradually decreased. The basic idea is that models with strong context utilization ability can compensate for missing information by relying on context. In the ``randomly mask one" setting, only one state in each sample was masked, while in the ``start \& end only" setting, only start and end states are provided. Figure~\ref{fig:crossmodal_miss} demonstrates TTNet has the highest robustness as more states are missing, highlighting its exceptional ability to utilize context for transformation reasoning. Comparing TTNet to two of its variants, one without MTM and one without semantic difference features, we concluded that both MTM and semantic difference features contribute to context utilization, with the latter having a greater impact.



\subsection{Analyses on Transformation Reasoning}  \label{sec:generalization}





\begin{table}[t]
\centering
\caption{Models including TTNet perform worse on unseen transformation combinations.}
\label{tab:res_combinations}
\small
\setlength{\tabcolsep}{0.3em}
\begin{tabular}{lcccccccc}
\toprule
& \multicolumn{4}{c}{Seen} & \multicolumn{4}{c}{Unseen}  \\
\cmidrule[.5pt](lr){2-5}  \cmidrule[.5pt](lr){6-9}
Model & C & Flu. & Rel. & Logic. & C & Flu. & Rel. & Logic. \\
\midrule
CST* & 0.99 & 1.95 & 3.22 & 3.00 & 0.73 & 2.17 & 3.08 & 2.91 \\
GLACNet* & 6.21 & 4.80 & 3.90 & 3.91 & 4.11 & 4.69 & 3.70 & 3.59 \\
DenseCap* & 5.16 & 4.72 & 3.66 & 3.61 & 3.75 & 4.76 & 3.68 & 3.57 \\
\midrule
TTNet$_\text{Base}$ & 6.02 & 4.80 & 4.08 & 4.00 & 4.40 & \textbf{4.77} & \textbf{3.99} & \textbf{3.88} \\
TTNet & \textbf{7.01} & \textbf{4.81} & \textbf{4.23} & \textbf{4.29} & \textbf{4.59} & 4.74 & 3.93 & 3.86 \\
\bottomrule
\end{tabular}
\end{table}

\textbf{Assessment on Reasoning Unseen Transformation Combinations.} A robust transformation reasoning system should be able to generalize to unseen transformation combinations, where individual transformations have been seen during training, but certain combinations have not. This often occurs when there are multiple ways of achieving the same task such as cooking noodles. In VTT, more than half of the combinations in the test set are not present in the training set (532 seen vs. 559 unseen). To evaluate how well models can reason about unseen transformation combinations, we divided the test set into two splits: ``seen" (combinations appeared in the training set) and ``unseen" (new combinations). As shown in Table~\ref{tab:res_combinations}, all models perform significantly worse on the unseen combinations than on the seen ones, with TTNet's logical soundness dropping by roughly 10\% (from 4.29 to 3.86), showcasing the challenge of generalization. The performance gap between TTNet, TTNet$_\text{Base}$, and DenseCap* on the unseen split is less significant than the gap on the seen split, implying that our strategies for modeling transformation primarily help with reasoning seen transformation combinations, while providing little benefit for reasoning unseen combinations.

\begin{figure}
\includegraphics[width=\linewidth]{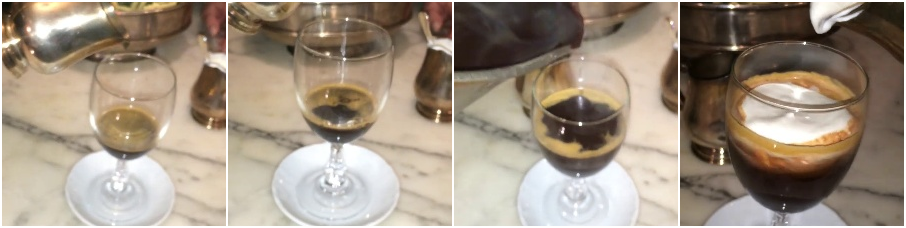}

\vspace{5pt}
\begin{tiny}

\begin{minipage}[c]{0.245\linewidth}

\textbf{DenseCap:} 

\textcolor{red}{1. Pour espresso.} 

\textcolor{red}{2. Pour espresso.} 

\textcolor{red}{3. Add whipped cream.} 

\end{minipage}
\begin{minipage}[c]{0.245\linewidth}

\textbf{GLACNet:} 

\textcolor{red}{1. Pour espresso.} 

\textcolor{red}{2. Pour espresso.} 

\textcolor{red}{3. Add whipped cream.} 

\end{minipage}
\begin{minipage}[c]{0.245\linewidth}

\textbf{TTNet:} 

\textcolor{red}{1. Pour alcohol.} 

\textcolor{red}{2. Pour espresso.} 

\textcolor{red}{3. Add whipped cream.} 

\end{minipage}
\begin{minipage}[c]{0.245\linewidth}

\textbf{Groundtruth:} 

1. Pour coffee into glass. 

2. Pour chocolate in glass. 

3. Pour cream. 
\end{minipage}
\end{tiny}

\caption{Models fail to describe unseen transformations composed by seen words.}
\label{fig:result_extra_example}
\end{figure}

\textbf{Assessment on Reasoning Unseen Language Compositions.} A robust transformation reasoning system should also be able to generalize to unseen language compositions, where individual words such as entities and actions have been seen during training, but their combinations have not. For example, successfully reasoning the unseen transformation ``pour coffee" when only ``pour milk" and ``make coffee" appeared in the training set. According to our statistics, VTT has a high proportion of shared vocabulary, this is the major reason that VTT is designed as a natural language generation task rather than a classification task, as models have a better chance of learning common patterns from transformations with shared words. To evaluate model generalization to new language compositions, we evaluated models on several manually labeled samples from ``related" tasks in CrossTask. In the example shown in Figure~\ref{fig:result_extra_example}, transformations for the topic \textit{Make Bicerin} have not appeared in VTT but are composed with seen words. However, all models failed to generate new descriptions and instead produced existing descriptions that matched the states as closely as possible. This indicates a significant limitation in the models' ability to generalize to new language compositions.

\subsection{Hyperparameter Tuning of MTM}

\begin{table}[t]
    \centering
    \caption{Results of different mask ratios used in MTM.}
    \label{tab:mask_ratio}
    \small
        \begin{tabular}{crrr}
            \toprule
            mask ratio & B@4            & C               & BS             \\
            \midrule
            0\%      & 60.38          & 562.83          & 75.72          \\
            5\%      & 60.93          & 567.92          & 76.11          \\
            10\%     & 61.02          & 568.71          & 76.13          \\
            15\%     & \textbf{61.22} & \textbf{570.63} & 76.25          \\
            20\%     & 61.07          & 568.99          & 76.21          \\
            25\%     & 61.16          & 570.18          & \textbf{76.35} \\
            30\%     & 60.72          & 565.43          & 75.94          \\
            \bottomrule
        \end{tabular}
\end{table}

\begin{table}[t]
    \centering
    \caption{Results of different sample ratios used in MTM.}
    \label{tab:mask_sample_ratio}
    \small
        \begin{tabular}{crrr}
            \toprule
            sample ratio & B@4            & C               & BS             \\
            \midrule
            0\%      & 60.38          & 562.83          & 75.72          \\
            25\%     & 60.39          & 562.15          & 75.63          \\
            50\%     & \textbf{61.22} & \textbf{570.63} & \textbf{76.25} \\
            75\%     & 60.96          & 567.99          & 76.00          \\
            100\%    & 60.95          & 568.18          & 76.10          \\
            \bottomrule
        \end{tabular}
\end{table}

There are two hyperparameters in the masked transformation model: the mask ratio and the sample ratio. The mask ratio is similar to that used in BERT~\cite{devlinBERTPretrainingDeep2019}, indicating the percentage of state representations and semantic difference features that are replaced with zero. After experimenting with mask ratios ranging from 0\%-30\%, we found 15\% works best (as shown in Table~\ref{tab:mask_ratio}), which is consistent with BERT's finding. The other hyperparameter is the sample ratio, which addresses the inconsistency between training and inference where no features are masked during inference. By setting the sample ratio, which is the probability that the sample will accept the masking strategy, we found a 50\% probability performs best, outperforming the strategy of masking all samples used in BERT (as shown in Table~\ref{tab:mask_ratio}).

\section{Comparison of Different Image Encoders} \label{sec:image_encoder}

The quality of image encoding is crucial for subsequent reasoning and description, which determines whether the model can correctly recognize and understand the image content. Therefore, image encoder significantly impacts the overall performance of the model. In the main paper, we observe that the original version of CST and GLACNet, with Inception V3~\cite{szegedyRethinkingInceptionArchitecture2016} and ResNet~\cite{heDeepResidualLearning2016} as image encoders, respectively, perform worse than CST* and GLACNet*. This indicates the importance of choosing an appropriate image encoder. We conduct a more detailed analysis of the image encoder by testing ten state-of-the-art image encoders, five of which were pretrained on ImageNet and five on large-scale image-text data from the CLIP variations. In the table, we report their parameter size, ImageNet top-1 accuracy, and performance on the VTT dataset. We found that when the parameter sizes were similar, models pretrained on image-text data outperformed those pretrained only on image data, e.g. ViT-L/14 vs. ViT-L. This is consistent with the existing understanding that CLIP encodes more semantic information. In addition to training data, factors that affect model performance include model size, patch size used in vision transformers, and training strategies. For example, CLIP models, which have more parameters, perform better. Although the parameter size between ViT-B/16 and ViT-B/32 is similar, ViT-B/16, which encodes finer images with smaller patch sizes, results in better image representation. BEiT-L~\cite{baoBEiTBERTPreTraining2022} has the highest accuracy on ImageNet but performs the worst among all models. We speculate that although BEiT-L has learned sufficient image pattern information, it has limitations in capturing semantic information.

\section{Additional Qualitative Results.}
\label{app:more_case}

\begin{figure}
\includegraphics[width=\linewidth]{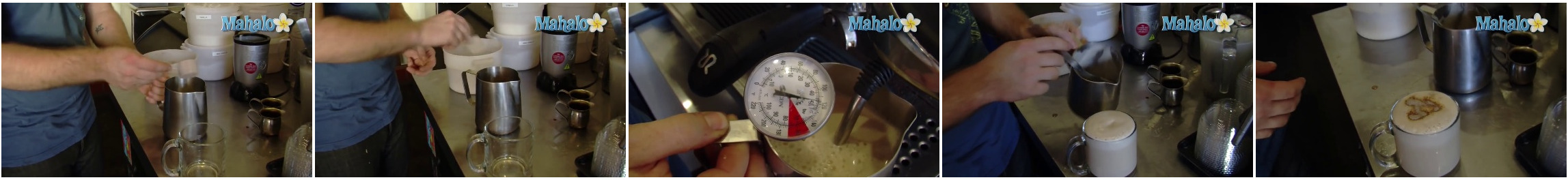}
\begin{tiny}

\vspace{2pt}

\setlength{\tabcolsep}{8pt}
\begin{tabular}{lrrrr}
\textbf{Groundtruth} 
& Add coffee.
& Steam milk.
& Pour milk.
& Add coffee.
\\

\textbf{Gemini-1.5}  
 & \textcolor{red}{Stir the milk}.
 & \textcolor{red}{Pour milk into mug}.
 & Top with froth.
 & Sprinkle cinnamon powder.
\\

\textbf{GPT-4o}
 & \textcolor{red}{Scoop ice into pitcher}.
 & \textcolor{red}{Add milk to pitcher.}
 & \textcolor{red}{Insert thermometer.}
 & \textcolor{red}{Use steam wand to heat milk.}
\\

\textbf{Qwen-VL-chat}
 & Man stirring coffee.
 & \textcolor{red}{Coffee being poured into mug.}
 & {\textcolor{red}Coffee being frothed.}
 & Coffee being decorated with heart design.
\\

\textbf{LLaVA-1.5-7b}
 & \textcolor{red}{Insert oil gun in the car.}
 & \textcolor{red}{Release liquid.}
 & \textcolor{red}{Put steak on grill.}
 & \textcolor{red}{Absorb liquid with dropper.}
\\

\textbf{TTNet}
 & \textcolor{red}{Pour milk.}
 & Steam milk.
 & Pour milk.
 & \textcolor{red}{Pour milk.}
\\

\end{tabular}
\end{tiny}
\vspace{6pt}

\includegraphics[width=\linewidth]{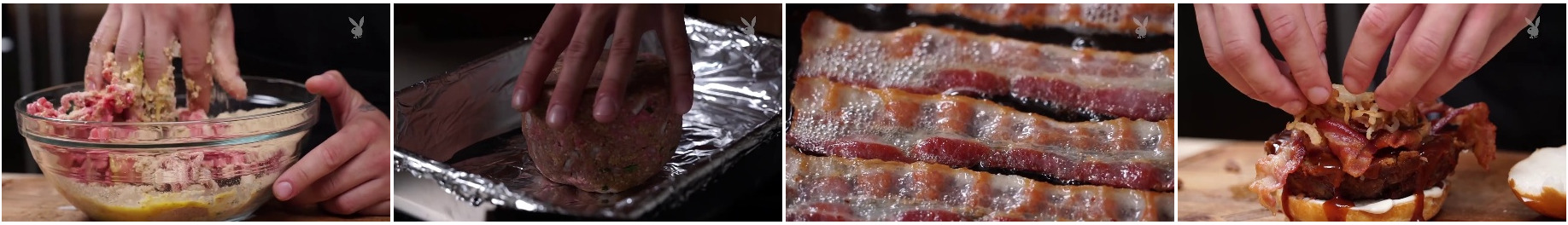}

\begin{tiny}

\vspace{2pt}

\setlength{\tabcolsep}{18pt}
\begin{tabular}{lrrr}
\textbf{Groundtruth}
& Knead the meat.
& Fry meat.
& Combine meat and bread to make burger.
\\

\textbf{Gemini-1.5}
 & Form meatloaf.
 & Fry bacon.
 & Assemble hamburger.
\\

\textbf{GPT-4o}
 & \textcolor{red}{Mix ingredients.}
 & \textcolor{red}{Form meatloaf.}
 & \textcolor{red}{Cook bacon.}
\\

\textbf{Qwen-VL-chat}
 & \textcolor{red}{Add bacon to burger.}
 & \textcolor{red}{Top with cheese.}
 & \textcolor{red}{Add caramelized onions.}
\\

\textbf{LLaVA-1.5-7B}
 & \textcolor{red}{Insert oil gun in the car.}
 & \textcolor{red}{Release liquid.}
 & \textcolor{red}{Put steak on grill.}
\\

\textbf{TTNet}
 & Knead the meat.
 & Fry meat.
 & Combine meat and bread to make burger.
\\
\end{tabular}
\end{tiny}
\vspace{6pt}

\includegraphics[width=\linewidth]{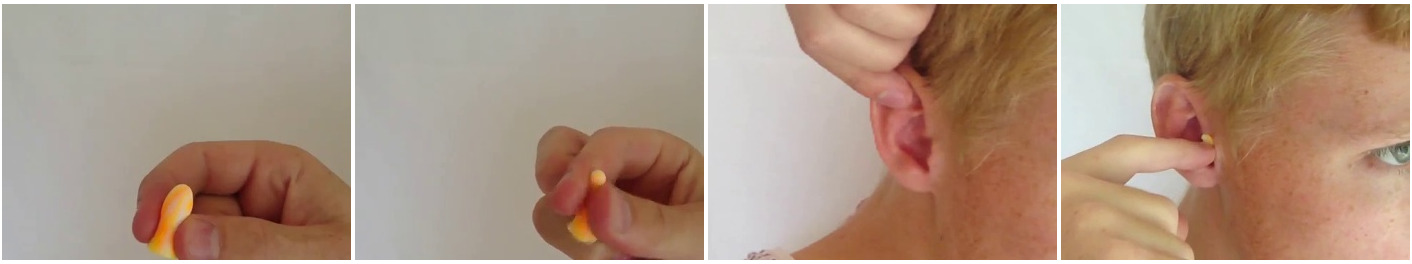}

\begin{tiny}

\vspace{2pt}

\setlength{\tabcolsep}{9pt}
\begin{tabular}{lrrr}
\textbf{Groundtruth}
& Roll and rub the earplugs.
& Pull the ears outwards.
& Insert the earplugs into the ears.
\\

\textbf{Gemini-1.5}
 & Roll earplug into a cylinder.
 & Place earplug in ear.
 & Push earplug into ear canal.
\\

\textbf{GPT-4o}
 & Hold earplug.
 & \textcolor{red}{Roll and compress earplug.}
 & \textcolor{red}{Pull back ear.}
\\

\textbf{Qwen-VL-chat}
 & \textcolor{red}{Remove candy corn from package.}
 & \textcolor{red}{Insert candy corn in ear.}
 & \textcolor{red}{Remove candy corn from ear.}
\\

\textbf{LLaVA-1.5-7B}
 & Hold orange object in hand.
 & \textcolor{red}{Insert orange object into the other hand.}
 & \textcolor{red}{Squeeze orange object to release its contents.}
\\

\textbf{TTNet}
 & Roll and rub the earplugs.
 & Pull the ears outwards.
 & Insert the earplugs into the ears.
\\
\end{tabular}
\end{tiny}

\vspace{6pt}

\includegraphics[width=\linewidth]{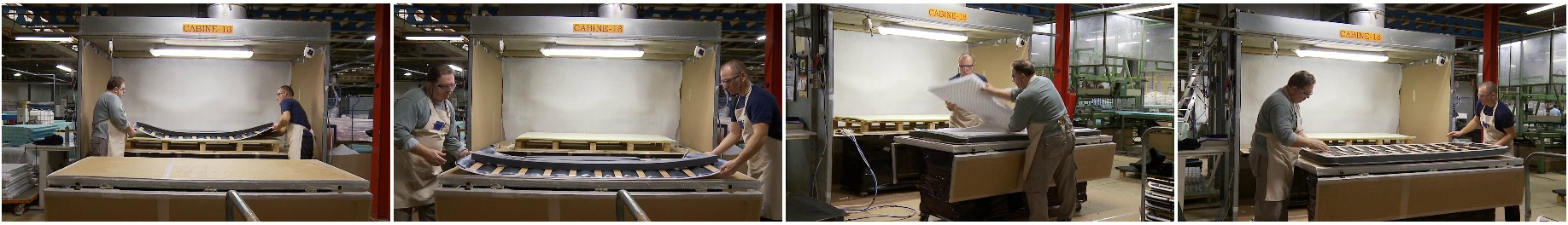}

\begin{tiny}

\vspace{2pt}

\setlength{\tabcolsep}{4pt}
\begin{tabular}{lrrr}
\textbf{Groundtruth}
& Place the bed board.
& Place mattress on the bed.
& Place mattress on the bed.
\\

\textbf{Gemini-1.5}
 & \textcolor{red}{Place padding over the board.}
 & Lay fabric over the padding.
 & \textcolor{red}{Remove the finished product.}
\\

\textbf{GPT-4o}
 & Remove frame from table.
 & \textcolor{red}{Place frame in workstation area.}
 & Adjust frame position.
\\

\textbf{Qwen-VL-chat}
 & \textcolor{red}{Man putting tray on machine.}
 & \textcolor{red}{Machine cutting material.}
 & \textcolor{red}{Man removing tray from machine.}
\\

\textbf{LLaVA-1.5-7B}
 & \textcolor{red}{Man wearing safety goggles.}
 & \textcolor{red}{Man wearing safety goggles and apron.}
 & \textcolor{red}{Man wearing safety goggles and apron while working on a project.}
\\

\textbf{TTNet}
 & Install stand of the bed.
 & \textcolor{red}{Install horizontal boards.}
 & \textcolor{red}{Install horizontal boards.}
\\
\end{tabular}
\end{tiny}

\vspace{6pt}

\includegraphics[width=\linewidth]{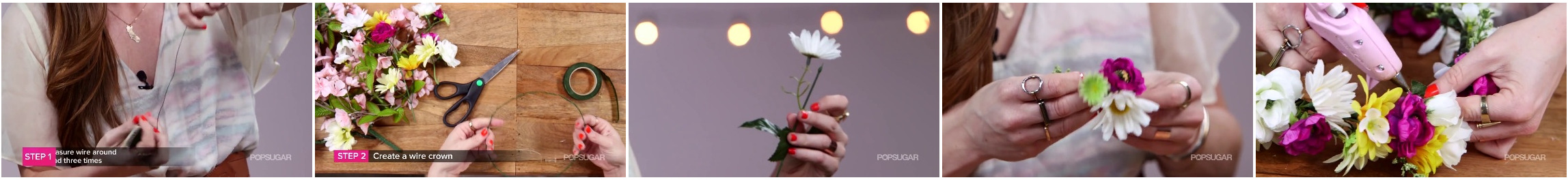}

\begin{tiny}

\vspace{2pt}

\setlength{\tabcolsep}{1pt}
\begin{tabular}{lrrrr}
\textbf{Groundtruth}
& Prepare the frame.
& Prepare flower.
& Prepare flower.
& Stick or bind flower to the frame.
\\

\textbf{Gemini-1.5}
 & \textcolor{red}{Wrap wire around hand three times.}
 & \textcolor{red}{Create a circular wire crown.}
 & Gather a cluster of flowers.
 & Attach flower cluster to wire crown.
\\

\textbf{GPT-4o}
 & \textcolor{red}{Measure wire around head three times.}
 & \textcolor{red}{Create a wire crown.}
 & \textcolor{red}{Cut flower stems.}
 & Attach flowers to crown using glue gun.
\\

\textbf{Qwen-VL-chat}
 & \textcolor{red}{Apply flower crown.}
 & Glue flowers together.
 & \textcolor{red}{Add greenery.}
 & \textcolor{red}{Finishing touches.}
\\

\textbf{LLaVA-1.5-7B}
 & \textcolor{red}{Flower petals wilt.}
 & \textcolor{red}{Flower petals dry.}
 & \textcolor{red}{Flower petals wither.}
 & \textcolor{red}{Flower petals die.}
\\

\textbf{TTNet}
 & Prepare the frame.
 & Prepare flower.
 & Stick or bind flower to the frame.
 & Stick or bind flower to the frame.
\\
\end{tabular}
\end{tiny}

\caption{More Cases of MLLMs and TTNet on the VTT test data. Error outputs are marked with \textcolor{red}{red}.}
\label{fig:more_cases}
\end{figure}

We present additional cases in Figure~\ref{fig:more_cases}.

\end{document}